%% file: main.tex
\documentclass[sigconf, nonacm, pdfa]{acmart}

\newcommand\vldbdoi{XX.XX/XXX.XX}
\newcommand\vldbpages{XXX-XXX}

\newcommand\vldbvolume{18}
\newcommand\vldbissue{11}
\newcommand\vldbyear{2025}
\newcommand\vldbauthors{\authors}
\newcommand\vldbtitle{\shorttitle} 
\newcommand\vldbavailabilityurl{https://github.com/zheng-yp/Lighter-X}
\newcommand\vldbpagestyle{empty} 

\usepackage{amsmath,amsfonts,dsfont,multirow,multicol,epsfig,url,array,makecell,balance,color,epstopdf}
\usepackage{algorithm}
\usepackage{algpseudocode}
\usepackage{amsmath}
\usepackage{hhline}
\usepackage{array}
\usepackage{subfigure}
\usepackage{booktabs}
\usepackage{xcolor,colortbl}
\usepackage{bm}
\usepackage{lipsum}
\usepackage{tabularx}
\usepackage{diagbox}
\usepackage[normalem]{ulem}
\useunder{\uline}{\ul}{}

\usepackage{enumerate}
\usepackage{enumitem}
\input{symbol-def}
\setlist[itemize]{leftmargin=*}
\newcommand{\eat}[1]{}

\usepackage[most]{tcolorbox}

\newcommand{\mymathinlinehl}[1]{\colorbox{gray!15}{$#1$}}
\newtcolorbox{mytextbox}
{colback=gray!10!white,colframe=black!75!black,boxsep=2pt,left=2pt,right=2pt,top=2pt,bottom=2pt}

\newcommand{\typosymbol}{\textsuperscript{\dag}}
\newcommand{\typofootnote}{\typosymbol~The \underline{\href{https://dcai-workshop.github.io/assets/pdf/accepted_papers/139.pdf}{workshop version}} contained minor typographical errors, which have been corrected in this revised submission.}

\begin{document}
\title{Lighter-X: An Efficient and Plug-and-play Strategy for Graph-based Recommendation through Decoupled Propagation}
\subtitle{[technical report]}

\author{Yanping Zheng}
\affiliation{%
  \institution{Renmin University of China}
  \country{}
}
\email{zhengyanping@ruc.edu.cn}

\author{Zhewei Wei}
\authornote{Zhewei Wei is the corresponding author. The work was partially done at Gaoling School of Artificial Intelligence, Beijing Key Laboratory of Research on Large Models and Intelligent Governance, MOE Key Lab of Data Engineering and Knowledge Engineering, Engineering Research Center of Next-Generation Intelligent Search and Recommendation, MOE, and Pazhou Laboratory (Huangpu), Guangzhou, Guangdong 510555, China.}
\affiliation{%
  \institution{Renmin University of China}
  \country{}
}
\email{zhewei@ruc.edu.cn}

\author{Frank de Hoog}
\affiliation{%
  \institution{Data 61, CSIRO}
  \country{}
}
\email{Frank.Dehoog@data61.csiro.au}



\author{Xu Chen}
\author{Hongteng Xu}
\affiliation{%
  \institution{Renmin University of China}
  \country{}
}
\email{xu.chen@ruc.edu.cn}
\email{hongtengxu@ruc.edu.cn}



\author{Yuhang Ye}
\author{Jiadeng Huang}
\affiliation{%
  \institution{Huawei Poisson Lab}
  \country{}
}
\email{yuhang.ye@huawei.com}
\email{huangjiadeng96@sina.com}

\input{sections/abstract}

\maketitle

\pagestyle{\vldbpagestyle}
\begingroup\small\noindent\raggedright\textbf{PVLDB Reference Format:}\\
\vldbauthors. \vldbtitle. PVLDB, \vldbvolume(\vldbissue): \vldbpages, \vldbyear.\\
\href{https://doi.org/\vldbdoi}{doi:\vldbdoi}
\endgroup
\begingroup
\renewcommand\thefootnote{}\footnote{\noindent
This work is licensed under the Creative Commons BY-NC-ND 4.0 International License. Visit \url{https://creativecommons.org/licenses/by-nc-nd/4.0/} to view a copy of this license. For any use beyond those covered by this license, obtain permission by emailing \href{mailto:info@vldb.org}{info@vldb.org}. Copyright is held by the owner/author(s). Publication rights licensed to the VLDB Endowment. \\
\raggedright Proceedings of the VLDB Endowment, Vol. \vldbvolume, No. \vldbissue\ %
ISSN 2150-8097. \\
\href{https://doi.org/\vldbdoi}{doi:\vldbdoi} \\
}\addtocounter{footnote}{-1}\endgroup

\ifdefempty{\vldbavailabilityurl}{}{
\vspace{.3cm}
\begingroup\small\noindent\raggedright\textbf{PVLDB Artifact Availability:}\\
The source code, data, and/or other artifacts have been made available at \url{\vldbavailabilityurl}.
\endgroup
}

\input{sections/intro}

\input{sections/pre}
\input{sections/investigation_lightgcn}

\input{sections/methods}
\input{sections/experiments}

\input{sections/experiments_otherscenarios}

\input{sections/conclusion}

\vspace{-2mm}
\begin{acks}
 This research was supported in part by National Natural Science Foundation of China (No. U2241212, No. 92470128), by Beijing Outstanding Young Scientist Program No.BJJWZYJH012019100020098, by Huawei-Renmin University joint program on Information Retrieval. We also wish to acknowledge the support provided by the fund for building world-class universities (disciplines) of Renmin University of China, by Engineering Research Center of Next-Generation Intelligent Search and Recommendation, Ministry of Education, by Intelligent Social Governance Interdisciplinary Platform, Major Innovation \& Planning Interdisciplinary Platform for the “Double-First Class” Initiative, Public Policy and Decision-making Research Lab, and Public Computing Cloud, Renmin University of China.
\end{acks}

\bibliographystyle{ACM-Reference-Format}
\bibliography{ref}

\clearpage
\appendix
\input{sections/appendix_notation}

\input{sections/appendix_model}
\input{sections/appendix_complexity}
\input{sections/appendix_otherscenarios}

\input{sections/appendix_exp}

\end{document}

%% file: symbol-def.tex
\newtheorem{observation}{Observation}

\def\p{\bm{p}}

\def\Z{\mathbf{Z}}

\def\I{\mathbf{I}}
\def\P{\mathbf{P}}
\def\A{\mathbf{A}}
\def\D{\mathbf{D}}
\def\R{\mathbf{R}}
\def\Q{\mathbf{Q}}
\def\S{\mathbf{S}}
\def\U{\mathbf{U}}
\def\V{\mathbf{V}}
\def\Y{\mathbf{Y}}
\def\X{\mathbf{X}}

\def\C{\mathbf{C}}
\def\W{\mathbf{W}}
\def\R{\mathbf{R}}
\def\B{\mathbf{B}}
\def\0{\mathbf{0}}
\def\J{\mathbf{J}}

\def\e{\bm{e}}
\def\p{\bm{p}}
\def\x{\bm{x}}
\def\y{\bm{y}}
\def\h{\bm{h}}
\def\E{\mathbf{E}}

\def\Dn{\mathbf{D}^{-\frac{1}{2}}}

\newcolumntype{C}[1]{>{\centering\arraybackslash}m{#1}} 

\definecolor{commentcolor}{RGB}{0, 100, 200} 
\definecolor{functioncolor}{RGB}{150, 50, 50}

\def\revision{}
\def\shepherd{}

%% file: sections/abstract.tex
\begin{abstract}
\label{abstract}
Graph Neural Networks (GNNs) have demonstrated remarkable effectiveness in recommendation systems. However, conventional graph-based recommenders, such as LightGCN, require maintaining embeddings of size $d$ for each node, resulting in a parameter complexity of $\mathcal{O}(n \times d)$, where $n$ represents the total number of users and items. This scaling pattern poses significant challenges for deployment on large-scale graphs encountered in real-world applications. To address this scalability limitation, we propose \textbf{Lighter-X}, an efficient and modular framework that can be seamlessly integrated with existing GNN-based recommender architectures. Our approach substantially reduces both parameter size and computational complexity while preserving the theoretical guarantees and empirical performance of the base models, thereby enabling practical deployment at scale. Specifically, we analyze the original structure and inherent redundancy in their parameters, identifying opportunities for optimization. Based on this insight, we propose an efficient compression scheme for the sparse adjacency structure and high-dimensional embedding matrices, achieving a parameter complexity of $\mathcal{O}(h \times d)$, where $h \ll n$. Furthermore, the model is optimized through a decoupled framework, reducing computational complexity during the training process and enhancing scalability. Extensive experiments demonstrate that Lighter-X achieves comparable performance to baseline models with significantly fewer parameters. In particular, on large-scale interaction graphs with millions of edges, we are able to attain even better results with only 1\% of the parameter over LightGCN.
\end{abstract}

%% file: sections/intro.tex
\section{Introduction}
\label{sec:intro}
\begin{figure}[t]
\setlength{\abovecaptionskip}{1mm}
\setlength{\belowcaptionskip}{2mm}
	\begin{small}
		\centering
		\begin{tabular}{cc}
		  \multicolumn{2}{c}{\includegraphics[height=3.2mm]{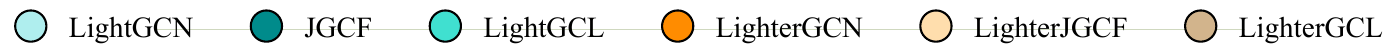}} \vspace{-1mm} \\
              \hspace{-5.3mm} \includegraphics[height=33mm]{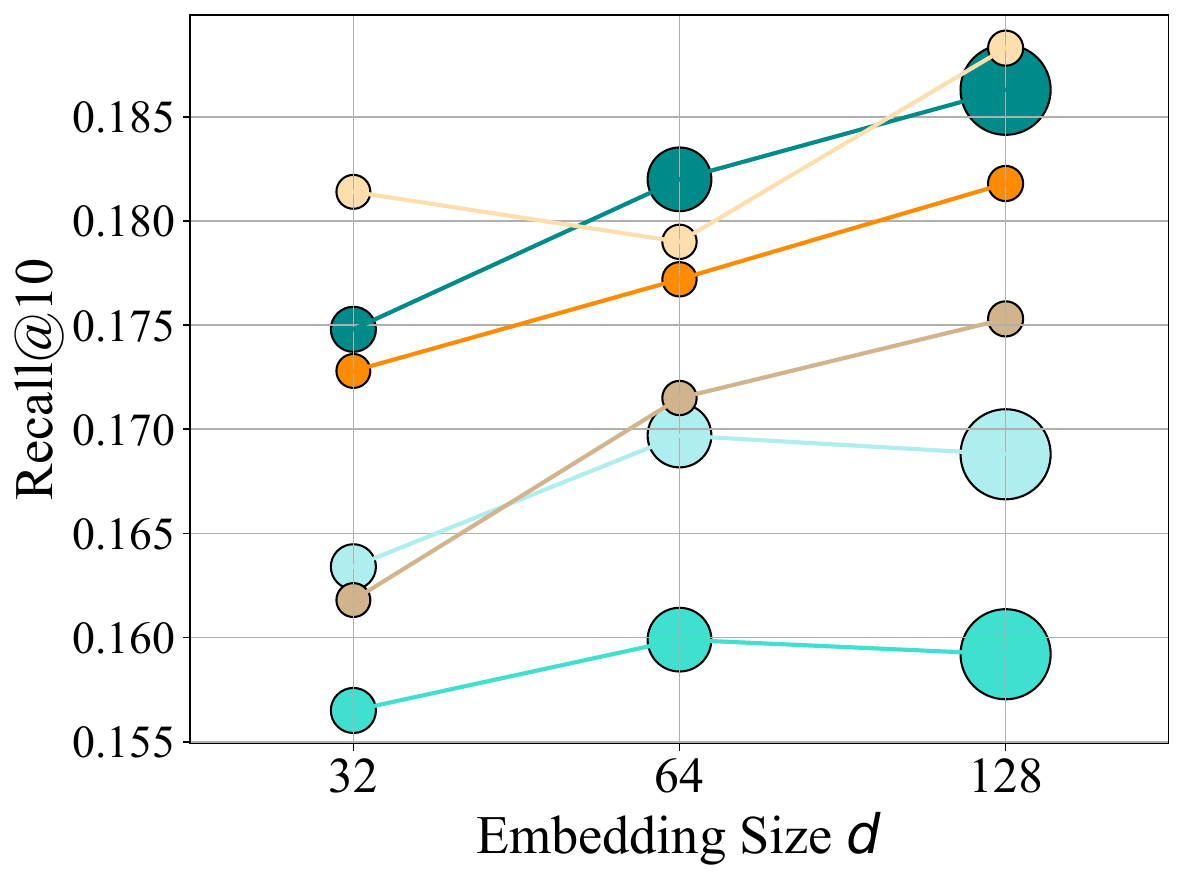} &
			 \hspace{-4.3mm} \includegraphics[height=33mm]{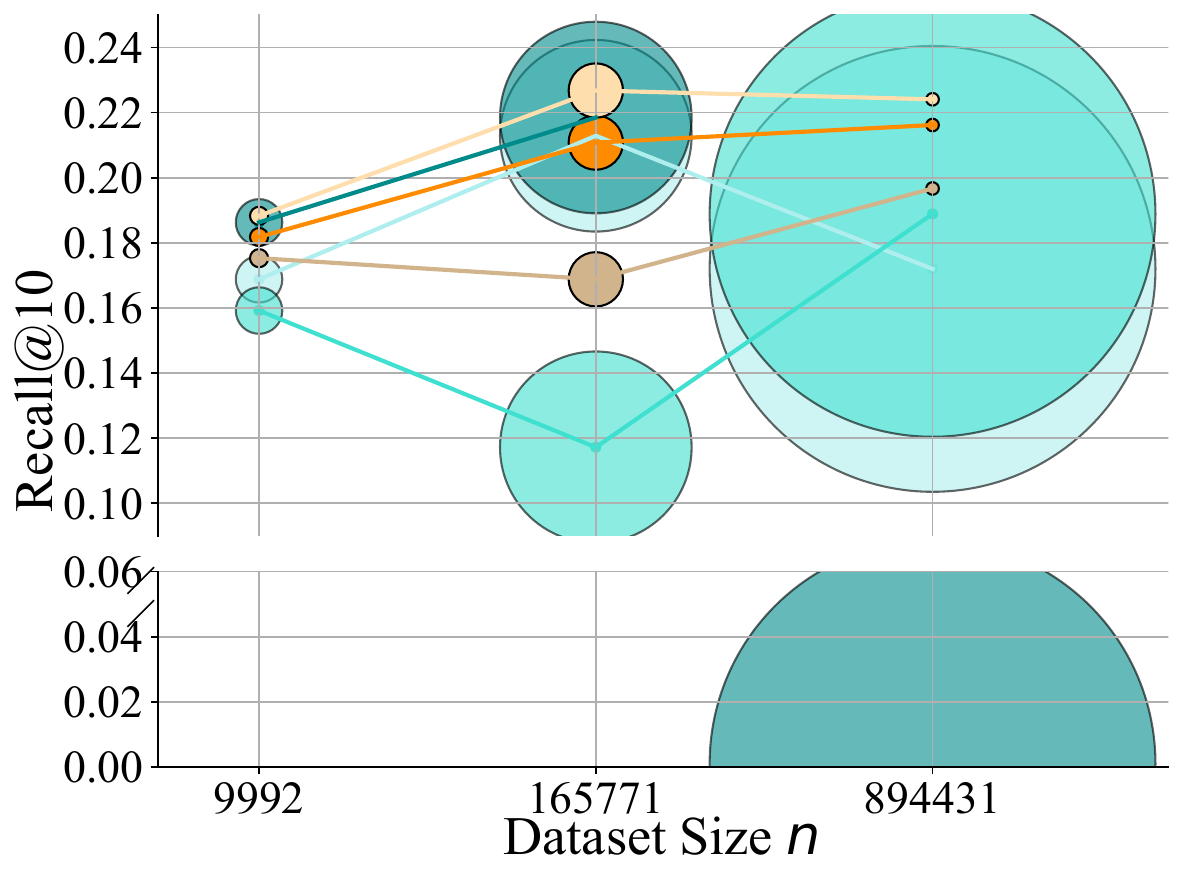} \\
		\end{tabular}
		\vspace{-2mm}
            \caption{Performance vs. Training Parameters: Circle sizes represent parameter counts. Baseline models' parameters scale proportionally with embedding size ($d$) and dataset size ($n$), while Lighter-X achieves higher accuracy with more compact parameter sizes.}
		\label{fig:params_muld_muln}
		\vspace{-6mm}
	\end{small}
\end{figure}
Recent studies have shown that recommender systems based on Graph Neural Networks (GNNs) outperform traditional collaborative filtering methods~\cite{wu2022survey_gnnrec}. Since much of the data in recommender systems can be naturally represented as graphs, GNNs leverage their powerful representation learning capabilities to capture complex relationships, thereby enhancing recommendation accuracy. For example, modeling user-item interactions as a bipartite graph allows for better exploitation of collaborative filtering information through neighbor convolution. 
By stacking more convolutional layers, the users and items with longer distances can be associated and share similar propagated gradients in the optimization process~\cite{gao2023survey_gnnrec}. Despite effectiveness, graph-based recommender models usually contain a large number of parameters and need complex convolutional operations, which hinders their application in real-world scenarios~\cite{adnan2021accelerating, peng2022svdgcn}. This problem necessitates the studies of more efficient graph-based recommender models. 

LightGCN~\cite{he2020lightgcn} simplifies traditional graph-based models by retaining only the essential neighbor aggregation operation. However, it still contains a large number of training parameters, expressed as $n \times d$, where $n$ is the total number of users and items, and $d$ is the embedding size. As shown in Figure~\ref{fig:params_muld_muln}, LightGCN's parameter count grows dramatically with both embedding dimension $d$ and dataset size $n$. Specifically, the left panel examines Recall@10 for varying embedding dimensions $d$ on the MovieLens-1M dataset. Overall, increasing $d$ leads to improved performance, but LightGCN~\cite{he2020lightgcn} requires significantly more parameters. The right shows results for models with fixed embedding dimensions across three datasets of increasing size: MovieLens-1M ($n$=9,992), MovieLens-20M ($n$=165,771), and Alimama ($n$=894,431). Similarly, LightGCN's parameter scales proportionally with dataset size $n$.

Recent works have introduced polynomial-based filters~\cite{guo2023jgcf} and Graph Contrastive Learning (GCL)~\cite{cai2023lightgcl, yu2022simgcl} to improve recommendation accuracy. However, these approaches rely on LightGCN~\cite{he2020lightgcn} as their backbone network, thereby inheriting its scalability limitations when applied to large-scale datasets, as shown in Figure~\ref{fig:params_muld_muln}. Notably, JGCF~\cite{guo2023jgcf} encountered an out-of-memory (OOM) error on the Alimama dataset. This raises an important question: \textbf{How can we design a lighter, more parameter-efficient framework while maintaining model performance?}

In this paper, we propose \textbf{Lighter-X}, a plug-and-play framework that can be seamlessly integrated into existing graph-based recommendation models to significantly reduce parameter cost. Motivated by the observation of inherent parameter redundancy in such models, we introduce a compression mechanism for both sparse graph structures and embedding matrices. 
As shown in Figure~\ref{fig:params_muld_muln}, Lighter-X models maintain stable model sizes regardless of embedding dimension $d$ or dataset size $n$, achieving parameter efficiency and competitive performance. 
Our contributions can be summarized as follows:
\begin{itemize}
\item We introduce Lighter-X, which reduces parameter complexity to $\mathcal{O}(h \times d)$, where $h\ll n$ corresponds to dataset sparsity.
\item Employing the Lighter-X framework, we improve existing recommender models and construct \textit{LighterGCN}, \textit{LighterJGCF} and \textit{LighterGCL}. Theoretical analysis shows that proposed models preserve the key properties of base models while significantly reducing parameter counts and computational complexity.
\item We conduct extensive experiments on several datasets and demonstrated that the proposed method achieves comparable or even better results with significantly fewer parameters, leading to substantially faster training times.
\end{itemize}

%% file: sections/pre.tex
\section{Background and Preliminary}
\label{sec:pre}
A recommender system typically consists of a user set $U$, an item set $I$, and a user-item interaction matrix $\R\in \{0, 1\}_{|U|\times|I|}$, where $\R_{ui}=1$ indicates an interaction between user $u$ and item $i$. Graph-based recommender models represent these interactions as a bipartite graph $G = (V, E)$, where the node set $V=U \cup I$ includes all users and items, and the edge set $E = \{(u, i) \mid \R_{ui} = 1, u \in U, i \in I\}$. 
The goal is to estimate user $u$'s preference for item $i\in I$ using their learned representation $\e_u$ and $\e_i$, formulated as $\hat{\y}_{u,i} = \e^\top_u \e_i$.

\vspace{-2mm}
\subsection{Decoupled GNNs}
GNNs are powerful tools for modeling graph data and have achieved impressive performance across various graph-related tasks. 
However, applying conventional GNNs such like GCN~\cite{kipf2016gcn} to large-scale graphs is challenging due to the limitations of full-batch training. To improve scalability without compromising accuracy, several methods, including SGC~\cite{wu2019sgc}, PPRGo~\cite{bojchevski2020pprgo}, and AGP~\cite{wang2021agp}, decoupled feature propagation from the training process. In general, feature propagation is computed as:
\begin{equation}
\label{equ:calZ}
\Z =\sum_{\ell=0}^{L} w_{\ell} \Z^{(\ell)}
 =\sum_{\ell=0}^{L} w_{\ell} \mathbf{\P}^\ell \X,
\end{equation}
where $L$ is the number of layers, $\P=\D^{-\frac{1}{2}} \A \D^{-\frac{1}{2}}$ is the normalized adjacency matrix, and $w_{\ell}$ denotes the importance of the $\ell$-th layer. Each $\Z^{(\ell)}$ s recursively defined as $\Z^{(\ell)} = \P\Z^{(\ell - 1)}$, with the initial representation $\Z^{(0)} = \X$, the input feature matrix (e.g., user attributes such as age, gender, or occupation). 
Typically, the feature propagation matrix $\Z$ can be precomputed and then used as input to a downstream model like a Multilayer Perceptron (MLP). In recommendation tasks, the goal is to learn node embeddings rather than prediction scores. With a single-layer MLP, the final embedding matrix is computed as $\E = \Z\W$, where $\W$ is the MLP weight matrix.

\vspace{-2mm}
\subsection{Graph-based Recommender Models}
\label{sec:pre_graph_based_models}
Graph-based recommender models learn powerful node embeddings by leveraging collaborative signals from high-order neighbors. NGCF~\cite{wang2019ngcf} is built on the standard GCN~\cite{kipf2016gcn} architecture. 
LightGCN~\cite{he2020lightgcn} simplifies NGCF by removing the weight matrices and the activation function in each layer. 
Formally, the embedding calculation in LightGCN can be represented by:
\begin{equation}
\label{equ:lightgcn}
\E=\frac{1}{L+1} \sum_{\ell=0}^L \E^{(\ell)}=\frac{1}{L+1} \sum_{\ell=0}^L \P^\ell \E^{(0)}, 
\end{equation}
where $L$ is the number of layers, $\E^{(\ell)}$ is the embedding matrix at layer $\ell$, and $\E^{(0)}$ is the initial embedding matrix, randomly initialized and used as the only learnable parameter. Each layer-wise embedding is computed recursively as $\E^{(\ell)}=\P \E^{(\ell-1)} = \P^\ell \E^{(0)}$. The repeated application of the propagation matrix $\P$ allows the model to capture multi-hop neighborhood information. Recent extensions introduce polynomial graph filters~\cite{guo2023jgcf} and graph contrastive learning~\cite{wu2021sgl, cai2023lightgcl, yu2022simgcl} to further boost performance.

\noindent \textbf{Polynomial graph filters.} Some works attribute the success of graph collaborative filtering to its effective implementation of low-pass filtering, and introduce polynomials to enable more flexible frequency responses~\cite{guo2023jgcf, qin2024polycf}. 
JGCF~\cite{guo2023jgcf} utilizes Jacobi polynomial bases, denoted as $\J_\ell^{a, b}(x)$, to approximate graph signal filters, facilitating efficient frequency decomposition and signal filtration. The $\ell$-th order Jacobi basis $\J_\ell^{a, b}(x)$ is parameterized by $a, b>-1$, which control the filter’s response characteristics. This formulation enables separate modeling of low- and mid-frequency signals, whose effects are combined to form the final embeddings:
\begin{equation}
\label{equ:jgcf}
\E=\text{concat}(\E_{low}, \E_{mid}), \quad \E_{low} = \frac{1}{L+1} \sum_{\ell=0}^L \J_\ell^{a, b}(\P) \E^{(0)}.
\end{equation}
The mid-frequency component is calculated as $\E_{mid}=\tanh(\beta\E^{(0)}-\E_{low}))$, where $\beta$ is a weighting factor controlling the balance between low- and high-frequency information.

\noindent \textbf{Graph contrastive learning. }To address the issue of sparse information in recommender systems, recent studies have introduced contrastive learning to enhance performance~\cite{cai2023lightgcl, wu2021sgl, yu2022simgcl}. The core idea is to modify the original graph structure to generate augmented representations. 
LightGCL~\cite{cai2023lightgcl} employs Singular Value Decomposition (SVD) to guide data augmentation. Specifically, SVD is applied to the interaction matrix $\R$, yielding $\R=\U\Q\V^\top$, where $\U \in \mathbb{R}^{|U| \times |U|}$ and $\V \in \mathbb{R}^{|I| \times |I|}$ are orthogonal matrices, 
and $\Q$ is a diagonal matrix of singular values. Since principal components correspond to top-$k$ singular values, LightGCL uses them to construct a perturbed interaction matrix $\hat{\R}$. The perturbed adjacency matrix $\hat{\A}=[[\0, \hat{\R}],[\hat{\R}^\top, \0]]$, which is then used in Equation~\ref{equ:lightgcn} to compute the perturbed embedding:
\begin{equation}
\hat{\E} = \sum_{\ell=0}^L \hat{\E}^{(\ell)}, \quad 
\hat{\E}^{(\ell)} = \hat{\P} \cdot \E^{(\ell-1)},
\end{equation}
where $\hat{\P} = \hat{\D}^{-\frac{1}{2}} \hat{\A} \hat{\D}^{-\frac{1}{2}}$ is the perturbed propagation matrix, $\hat{\E}^{(\ell)}$ refers to the perturbed embedding at layer $\ell$, and $\hat{\E}^{(0)} = \E^{(0)}$. 

{\revision
\noindent \textbf{Scalable methods.} To improve the scalability of graph-based recommendation systems, several approaches have been proposed to balance efficiency and memory use. XGCN~\cite{song2024xgcn} is a library designed for GNN-based recommendations, incorporating optimized implementations and scaling strategies to process large datasets with low memory overhead. LTGNN~\cite{zhang2024ltgnn} enhances propagation efficiency by adopting an implicit modeling approach inspired by PPNP and integrating a variance-reduced neighbor sampling strategy to further improve scalability and efficiency. GraphHash~\cite{wu2024graphhash} focuses on parameter reduction by employing modularity-based bipartite graph clustering to compress the embedding table. This approach is orthogonal to our work, as Lighter-X improves parameter efficiency by optimizing the model’s computational structure.

\noindent \textbf{Simplified methods.} Recently, some works have been proposed to optimize and simplify graph-based recommendation models. LightGODE~\cite{zhang2024lightgode} reduces training cost by modeling graph convolution as differential equations, removing graph operations during training and reintroducing them only for validation.  However, its structure remains similar to LightGCN, with no reduction in parameters. Another line of work, such as SVD-GCN~\cite{peng2022svdgcn}, reduce parameters via truncated SVD for low-rank embedding approximation. While effective, SVD incurs high time and memory costs on large-scale graphs, limiting scalability. In contrast, the proposed Lighter-X achieves both computational simplification and parameter compression.
}

%% file: sections/investigation_lightgcn.tex
\vspace{-2mm}
\section{Investigation of Graph-Based Recommendation Models}
\label{sec:investigation_lightgcn}
In this section, we analyze the connection between LightGCN~\cite{he2020lightgcn} and decoupled GNN models, highlighting the reasons behind the large parameter sizes in graph-based recommendation models, using LightGCN as a representative example. We then demonstrate through experimental observations that this large parameter matrix is largely redundant.
\noindent \textbf{Origins for Large Parameter Counts.} LightGCN~\cite{he2020lightgcn} simplifies NGCF~\cite{wang2019ngcf} by removing feature transformations and nonlinear activations, relying solely on linear neighborhood aggregation to capture collaborative signals. It can be viewed as a simplified form of a decoupled GNN. While LightGCN is not fully decoupled, since it still aggregates node representations at each layer, it behaves equivalently to a decoupled GNN in terms of parameterization and embedding learning. This equivalence can be demonstrated by setting $w_\ell = 1/(L+1)$ in Equation~\ref{equ:calZ} and letting $\X=\I$, where $\I$ is the $n\times n$ identity matrix. Under these settings, Equation~\ref{equ:calZ} becomes $\Z=1/(L+1)\sum_{\ell=0}^L \P^\ell \I$. Substituting this into the embedding computation yields $\mymathinlinehl{\E=\Z \W=1/(L+1) \sum_{\ell=0}^L \P^\ell \I \W}$, which matches Equation~\ref{equ:lightgcn}, where $\E^{(0)}$ corresponds to the parameter matrix $\W$ in decoupled GNNs. This equivalence is further supported by empirical results presented in our technical report\eat{~\cite{technical_report}}. Observation~\ref{observation:key} aligns perfectly with the statement in recommender systems that \textbf{the IDs of users and items are used as input features}. In these systems, users and items lack intrinsic features beyond their IDs, which effectively results in a one-hot encoded input. This setup is analogous to a scenario in decoupled GNNs where an identity matrix serves as the feature matrix.
\vspace{-1mm}
\begin{mytextbox}
\begin{observation}
\label{observation:key}
In terms of embedding learning and model parameters, LightGCN can be seen as a specialized form of decoupled GNN, where the input feature matrix is set to an identity matrix.
\end{observation}
\end{mytextbox}
\vspace{-1mm}

According to the mathematical formulation, the dimensions of the parameter matrix $\W$ are determined by the feature dimensions. When $\X$ is an identity matrix, the feature dimension becomes $n$, resulting in a parameter size of $n\times d$ for LightGCN~\cite{he2020lightgcn}. JGCF~\cite{guo2023jgcf} and LightGCL~\cite{cai2023lightgcl} employ polynomial-based filters and GCL, respectively, to improve model performance. Due to their adherence to LightGCN's embedding learning framework, their large parameter sizes can be attributed to the same factors outlined previously.


\begin{figure}[t]
\setlength{\abovecaptionskip}{1mm}
\setlength{\belowcaptionskip}{-1mm}
	\begin{small}
		\centering
		\vspace{-1mm}
		\begin{tabular}{c}
			 LastFM \\
			 \hspace{-4.3mm} \includegraphics[height=23mm]{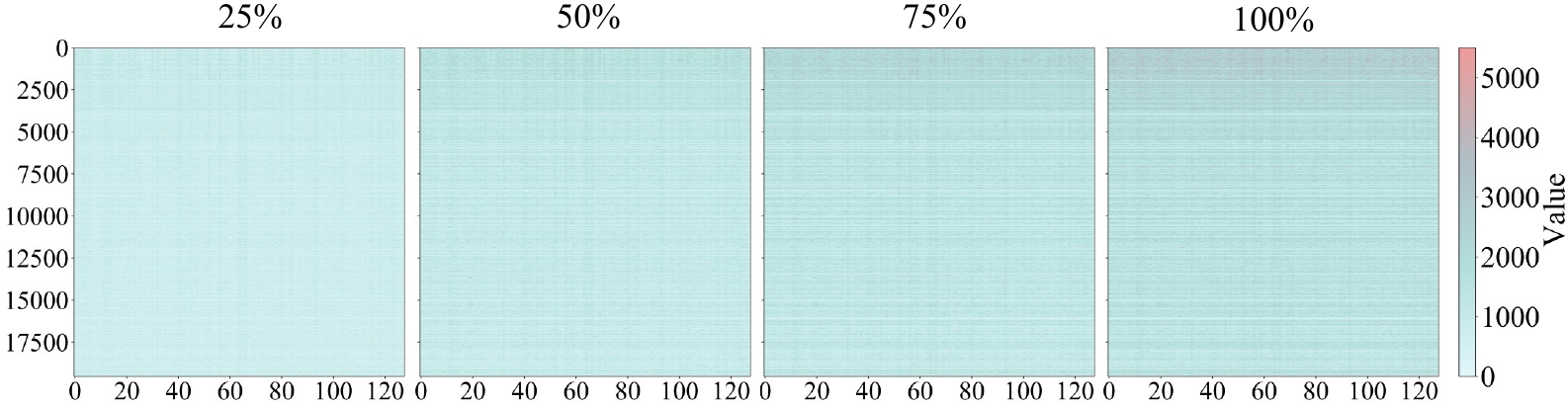} \\ 
			 MovieLens-1M \\
			 \hspace{-4.3mm} \includegraphics[height=21.8mm]{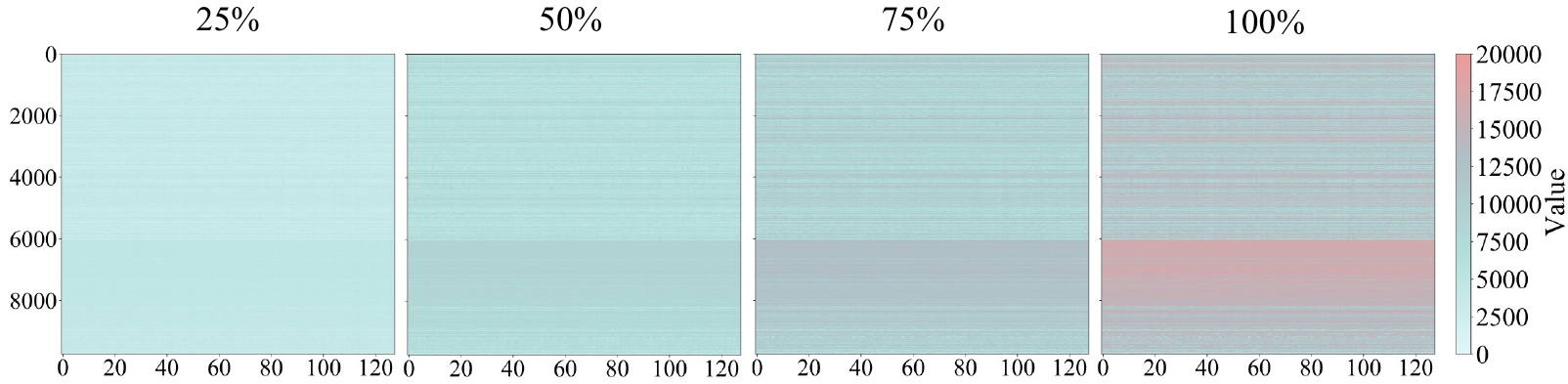} \\
		\end{tabular}
		\vspace{-2mm}
		\caption{Parameter matrix updates during training.}
		\label{fig:updated_times_tensor_W}
		\vspace{-4mm}
	\end{small}
\end{figure}
\begin{figure}[t]
\setlength{\abovecaptionskip}{1mm}
\setlength{\belowcaptionskip}{-1mm}
	\begin{small}
		\centering
		\begin{tabular}{cc}
			 \hspace{-4.3mm} \includegraphics[height=25mm]{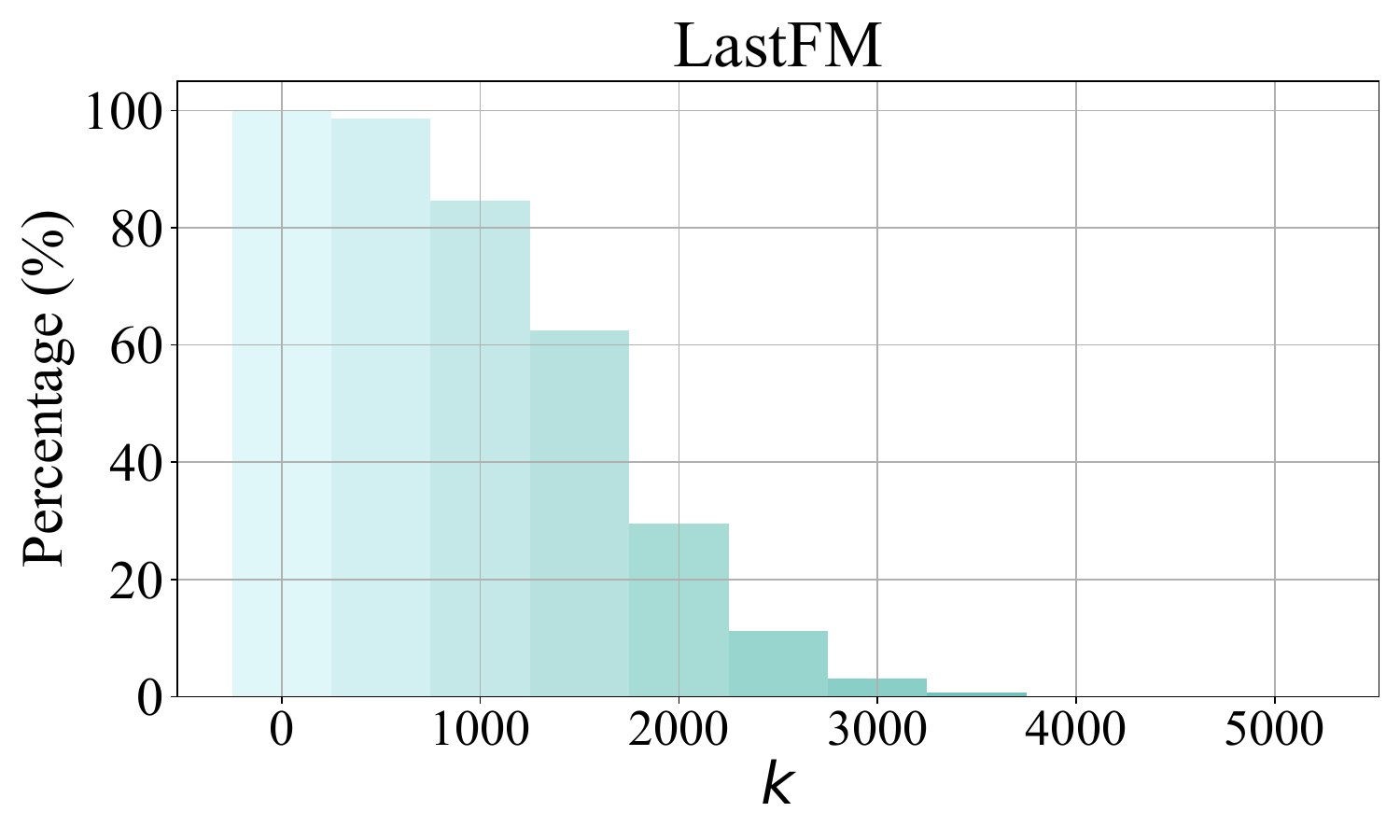} &
			 \hspace{-3.3mm} \includegraphics[height=25mm]{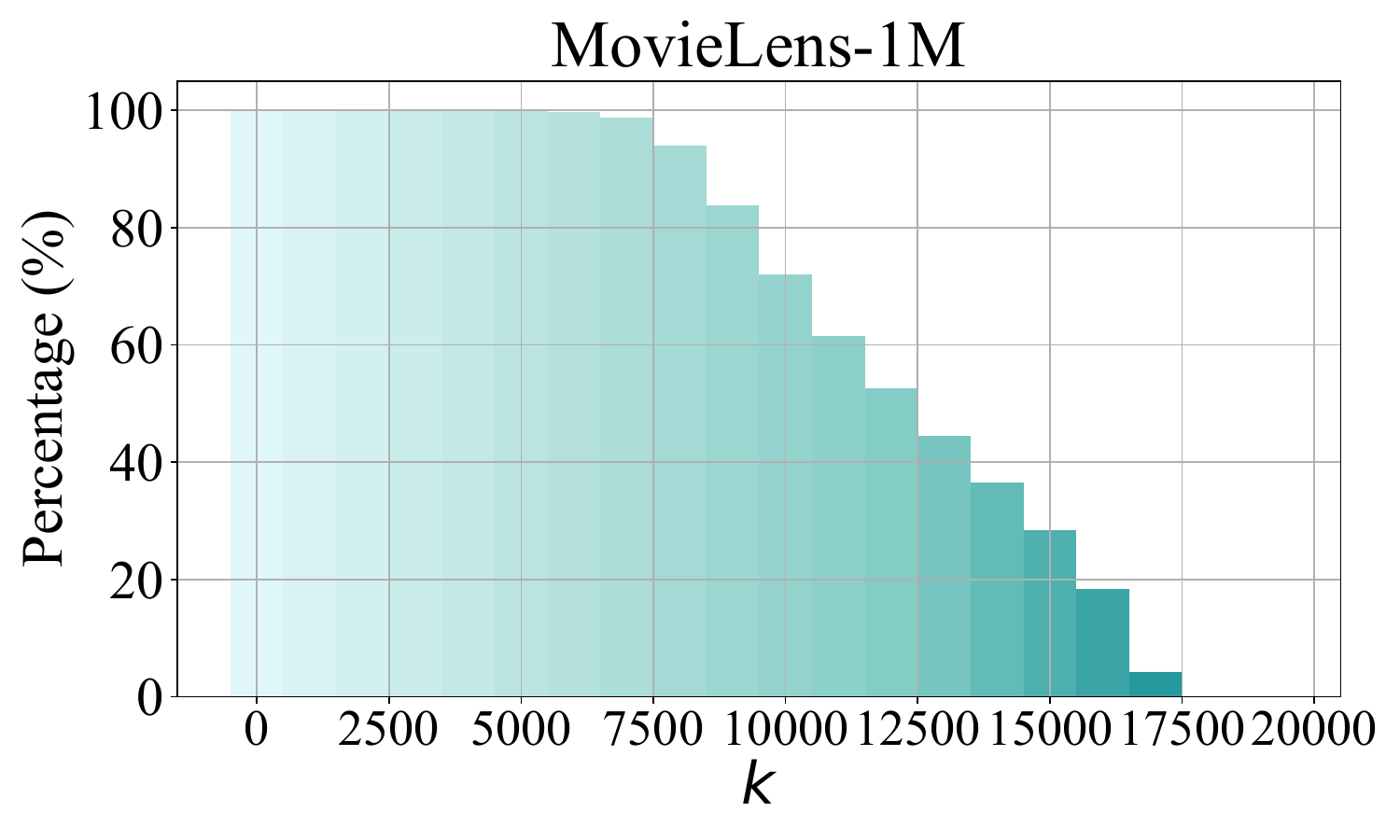} \\
		\end{tabular}
		\vspace{-2mm}
		\caption{Percentage of parameter updated more than $k$ times.}
		\label{fig:updated_times_W}
		\vspace{-5mm}
	\end{small}
\end{figure}

\noindent \textbf{Redundancy in Parameter Matrices.} Considering that the parameter matrix in LightGCN~\cite{he2020lightgcn} and its variants scales with $ n \times d $, we conducted experiments on the LastFM and MovieLens-1M datasets to examine its necessity and potential redundancy. Specifically, we tracked parameter update frequencies during training to assess utilization. As shown in Figure~\ref{fig:updated_times_tensor_W}, only a small portion of the parameter matrix continues to update during the later stages of training. This effect is particularly evident on the LastFM dataset, where many parameters become static in the early phases, indicating redundancy. To further investigate this, Figure~\ref{fig:updated_times_W} shows the percentage of parameters updated more than $k$ times. On MovieLens-1M, most parameters are updated infrequently, and the trend is even more pronounced in the LastFM dataset, where fewer than 20\% of the parameters are updated more than 2,500 times out of a possible 5,000. These findings suggest that the parameter matrix $\W$, also referred to as $\E^{(0)}$, in LightGCN is highly redundant. Since many graph-based recommender models adopt similar parameter settings, this highlights a broader need for parameter optimization. 

\begin{mytextbox}
\begin{observation}
\label{observation:parameters}
Parameter matrices in models like LightGCN exhibit significant redundancies, demonstrating that the training parameter matrix is inherently sparse.
\end{observation}
\end{mytextbox}
To address this, we propose a foundational assumption: $\W = \W_\text{s} + \W_\text{v}$, where $\W_\text{s}$ consists of static parameters and $\W_\text{v}$ contains the learnable, varying components. 
Correlating the results from Figures~\ref{fig:updated_times_tensor_W} and~\ref{fig:updated_times_W}, it becomes evident that $\W_\text{v}$ should be a sparse matrix. The redundancy observed in the parameter matrix suggests that a large portion of $\W$ remains effectively unchanged during training and does not significantly contribute to the learning process. Our empirical results confirm that only a small subset of parameters in $\W_\text{v}$ are meaningfully updated, highlighting a clear opportunity for reducing model size and improving efficiency.

%% file: sections/methods.tex
\section{The Lighter-X Method}
\label{sec:lighterx}
In this section, we introduce the \textbf{Lighter-X} framework and demonstrate its universal applicability by applying it to various representative models. As discussed in Section~\ref{sec:investigation_lightgcn}, LightGCN uses an identity matrix of dimension $\mymathinlinehl{n \times n}$ as the feature matrix $\X$, which necessitates a weight matrix $\W$ of size $\mymathinlinehl{n \times d}$. This setup results in a substantial number of parameters. To tackle this issue, we propose using a low-rank matrix $\X \hspace{-0.5mm}\in\hspace{-0.5mm} \mathbb{R}^{n \times h}$ as the input feature matrix. This adjustment results in a weight matrix $\W$ of size $\mymathinlinehl{h \times d}$ according to the standard computation, where $\mymathinlinehl{h \ll n}$. \textbf{Given this context, what constitutes an optimal low-rank matrix $\X$?}

\begin{figure}[t]
\centering
\setlength{\abovecaptionskip}{1mm}
\setlength{\belowcaptionskip}{-1mm}
\hspace{-2mm}\includegraphics[height=28mm]{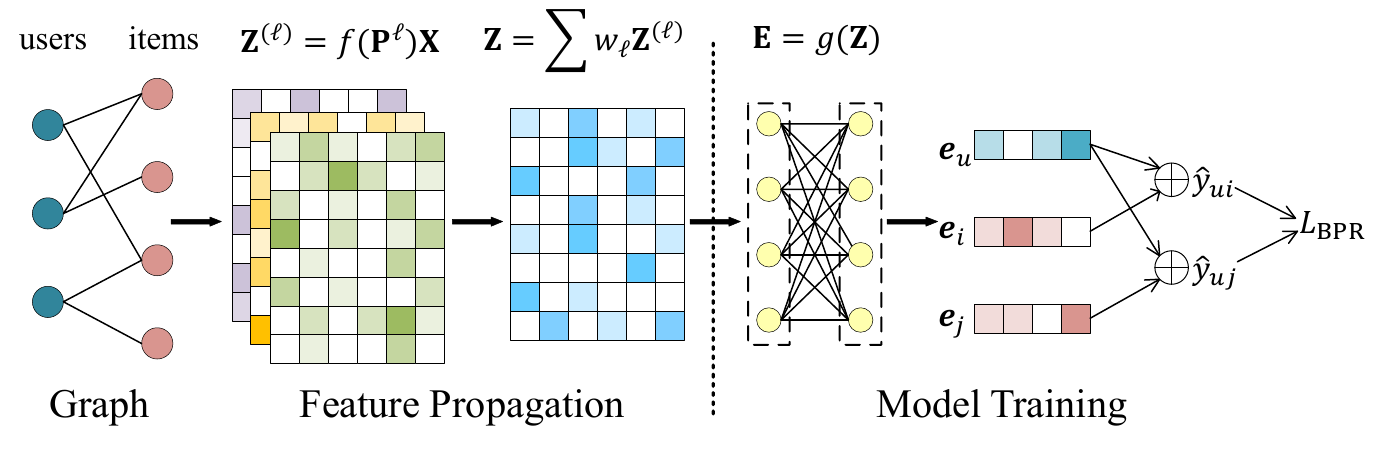}
\caption{An overview of the proposed Lighter-X framework.}
\vspace{-2mm}
\label{fig:decoupled}
\end{figure}
In recommender systems, data is typically large-scale but inherently sparse. As discussed in Section~\ref{sec:investigation_lightgcn}, this sparsity extends to the training parameters of the models. Building on this characteristic, our approach leverages compressed sensing to efficiently derive low-rank matrices, which is essential for managing large data volumes with reduced computational overhead. This method provides a robust alternative to traditional techniques such as SVD. Although popular for achieving low-rank approximations, SVD is challenged by significant computational demands in large graph scenarios~\cite{du2023treesvd}. 
To achieve the restricted isometry property (RIP) in the optimal regime for compressed sensing, we utilize random matrices, a common technique for rapid dimensionality reduction. Even under significant compression, the original signal can be accurately reconstructed from a small number of observations, provided the signal retains sparse characteristics. This reconstruction is achieved through optimization algorithms~\cite{foucart2013csbook}. In other words, the compressed matrix can preserve the essential features of the data, making compressed sensing a promising approach for accelerating convolutional operations by effectively reducing dimensionality early in the network.

{\revision 
\noindent \textbf{Optimizing sparse data in graph structures.} To optimize the sparse data in graph structures, we construct an efficient input feature matrix $\X = \P \cdot \S$ using cost-effective random sampling, where $\S \in \mathbb{R}^{n \times h}$ is a random matrix designed to satisfy the RIP condition. Specifically, $\S$ can be either a Gaussian or Bernoulli random matrix, both of which are widely used in compressed sensing due to their simplicity, generality, and ability to satisfy the RIP condition~\cite{yang2014gaussian, do2008gaussianbernoulli, lu2019bernoulli}. The dimensions of $\S$ are chosen to meet the following requirement:
\begin{equation}
\label{equ:h_dimension}
h = c \cdot r \log(n/r) ,
\end{equation}
where $r$ represents the sparsity level and $c$ is a customizable constant. Traditional methods maintain graph propagation precision by generating an ID-specific one-hot vector for each node, which leads to inefficient resource usage, as this approach requires $n$ entries for $n$ signals. In contrast, by utilizing compressed sensing and random sampling, as described in Equation~\ref{equ:h_dimension}, our method scales with $\log(n)$, significantly reducing resource consumption while preserving essential features. }

{\revision
\noindent \textbf{Optimizing sparse trainable parameters.} Following a similar strategy used for sparse graph structures, we replace the sparse parameter matrix discussed in Section~\ref{sec:investigation_lightgcn} with a trainable matrix $\W^\prime$, initialized from a Gaussian distribution. Since the random projection matrix $\S^\prime$ is also sampled from a Gaussian distribution~\cite{yang2014gaussian}, their product $\S^\prime \W^\prime$ satisfies the distributional properties required in compressed sensing~\cite{do2008gaussianbernoulli}. Importantly, the dimensionality of learnable weight $\W^\prime$ is $\mymathinlinehl{h \times d$, independent of the number of nodes $n}$, which contributes to improved scalability. Therefore, the model can bypass the traditional reconstruction step and instead rely on end-to-end training to learn effective representations. 
}

\noindent \textbf{Decoupled framework for graph-based recommendation.} The coupled model structure is another important factor limiting the scalability of traditional GCN~\cite{kipf2016gcn} and LightGCN~\cite{he2020lightgcn}. Specifically, these models typically require convolutional operations to be performed on the entire graph, which is computationally expensive and difficult to scale to large graphs. A series of studies has improved GCN scalability by decoupling feature propagation from the training process, allowing computationally intensive convolution operations to be precomputed~\cite{wu2019sgc, bojchevski2020pprgo, wang2021agp}. Extending this idea, our introduction of low-rank random matrices enables the decoupling of Lighter-X, allowing the costly and time-consuming feature propagation operations to be executed only once during the pre-computation phase. \eat{Subsequent training for the downstream tasks can utilize a standalone parameter matrix, as shown in Equation~\ref{equ:lighter_W}, or a slightly more complex MLP. }Figure~\ref{fig:decoupled} illustrates the final Lighter-X framework, where $f(\cdot)$ is the propagation function responsible for spreading information across nodes, and $g(\cdot)$ is the learning function, typically implemented as an MLP trained for downstream tasks. In the feature propagation stage, we complete the convolution related operation and obtain the feature propagation matrix $\Z$. The subsequent neural network takes $\Z$ as input and is trained to generate the final user and item embeddings. This training process is guided by the Bayesian Personalized Ranking (BPR) loss. 



\begin{algorithm}[t]
    \caption{\shepherd Training Algorithm for LighterGCN}
    \label{alg:training}
    \begin{algorithmic}[1]
        \State \textbf{Input:} User-item interaction matrix $\R$, adjacency matrix $\A$, degree matrix $\D$, number of GNN layers $L$, random matrix dimension coefficients $c$
        \State \textbf{Output:} Predicted score matrix $\hat{\Y}$, learned embeddings $\E$
        
        \State \textcolor{commentcolor}{\textit{\# Preprocessing}}
        \State Compute normalized adjacency matrix $\P = \D^{-\frac{1}{2}} \A \D^{-\frac{1}{2}}$
        \State Generate feature matrix $\X = \textcolor{functioncolor}{\textsc{GenFeat}}(\R, c)$
        \State Compute feature propagation matrix $\Z=\sum_{\ell=0}^{L} w_\ell \P^\ell \X$

        \State \textcolor{commentcolor}{\textit{\# Training}}
        \For{each mini-batch with $B$ user-item pairs $(u, i, i^{-})$}
            \State $\Z_B = \text{rows of }\Z \text{ indexed by } \{u, i, i^{-}\}$
            \State Get embeddings for nodes in batch $\E_B = MLP(\Z_B)$

            \State $\mathcal{L}_{\mathrm{BPR}}=-\log \left[\operatorname{sigmoid}\left(\boldsymbol{e}_u^{\top} \boldsymbol{e}_i-\boldsymbol{e}_u^{\top} \boldsymbol{e}_{i^{-}}\right)\right]$
            \State Update MLP's parameters using gradient descent
        \EndFor

        \State \textcolor{commentcolor}{\textit{\# Inference}}
        \State Get embeddings for all nodes $\E = MLP(\Z)$
        \State $\E_U = \text{rows of }\E \text{ indexed by } \{1, \dots, |U|\}$
        \State $\E_I = \text{rows of }\E \text{ indexed by } \{|U|+1, \dots, |U|+|I|\}$
        \State Predict score matrix $\hat{\Y} = \E_U \E_I^{\top}$


    \end{algorithmic}
\end{algorithm}

\vspace{-2mm}
\subsection{LighterGCN}
\label{sec:lightergcn}
We begin by applying the proposed Lighter-X framework to extend LightGCN~\cite{he2020lightgcn}, which we call LighterGCN. This is particularly relevant, since LightGCN serves as an foundational backbone for many GNN-based recommendation models. Specifically, LighterGCN adopts a low-rank approximation and decoupling framework to optimize the embedding process. Formally, LighterGCN learns embeddings using the following equation:
\begin{equation}
\label{equ:lightergcn}
\vspace{-0.5mm}
\E = MLP(\Z) = MLP(\sum_{\ell=0}^L w_\ell \Z^{(\ell)}), \quad 
\Z^{(\ell)} = \P^\ell \X,
\vspace{-1mm}
\end{equation}
where $\X$ is the random sampling result with rank $h$, which is much smaller than the number of nodes $n$. Based on this low-rank input feature matrix $\X$, LighterGCN performs graph convolutional operations to compute the feature propagation matrix $\Z$. Finally, an MLP is trained to produce the final embedding $\E$. As a result, LighterGCN reduces the number of parameters from $O(nd)$ to $O(hd)$, where $h \ll n$, thereby simplifying computation and improving learning efficiency. By precomputing feature propagation using the introduced low-rank random matrix, LighterGCN not only maintains the expressive power of the original LightGCN but also achieves greater scalability and efficiency. 

{\shepherd
\noindent \noindent \textbf{Learning Algorithm}. The LighterGCN method, summarized in Algorithm~\ref{alg:training}, consists of three main stages: preprocessing, training, and inference. During preprocessing (Lines 4–6), we first compute the normalized adjacency matrix, as is standard in many existing methods. We then generate feature matrices using a randomized approach and construct the feature propagation matrix following the LightGCN formulation, using the low-rank feature matrix as input. This shared propagation mechanism enables LighterGCN to effectively preserve the strengths of LightGCN. In the training phase (Lines 8–12), we sample mini-batches of user-item pairs and learn embeddings using an MLP. Importantly, no graph-related operations are required during training, which significantly improves efficiency. Since each row of the feature propagation matrix is independent, computations are restricted to the relevant nodes in each mini-batch, avoiding redundant full-graph convolutions and further enhancing scalability. Finally, during inference (Lines 15–18), we compute predicted user-item relevance scores by multiplying the learned embeddings. To facilitate understanding and comparison of computational stages, we present an overview of the training pipeline. Further details are available in the technical report\eat{~\cite{technical_report}}.
}

\begin{table*}[t]
\caption{The comparison of time complexity between baseline and proposed models. $n$, $m$, $|U|$ and $|I|$ represent the number of nodes, edges, users and items, respectively. $B$ represents the batch size, $n_B$ denotes the number of nodes in a batch, $L$ is the number of layers in the model, $d$ refers to the embedding size, $h$ is the dimension of the feature matrix, and $q$ is the required rank. $T$ denotes the number of iterations in training and is equal to $m/ B$.}
\label{tab:complexity}
\setlength{\abovecaptionskip}{1mm}
\setlength{\belowcaptionskip}{-4mm}
\centering
\begin{tabular}{cC{1.2cm}C{1.8cm}|C{1.4cm}|C{1.4cm}|C{2.2cm}|C{1.6cm}|C{1.7cm}|C{2cm}}
\Xhline{1.2pt} 
\multicolumn{2}{c|}{\textbf{Stage}}                                                                                                           & \textbf{Computation}                                         & \textbf{LightGCN} & \textbf{JGCF} & \textbf{LightGCL}    & \textbf{LighterGCN}             & \textbf{LighterJGCF}            & \textbf{LighterGCL}               \\ \hline
\multicolumn{2}{c|}{\multirow{3}{*}{\begin{tabular}[c]{@{}c@{}}\textbf{Pre-processing}\end{tabular}}}                                             & Normalization                                                & $O(2m)$         & $O(2m)$         & $O(2m)$                & $O(2m)$          & $O(2m)$          & $O(2m)$            \\ \cline{3-9} 
\multicolumn{2}{c|}{}                                                                                                                       & SVD                                                          & -             & -             & $O(qm)$                & -              & -              & $O(qm)$            \\ \cline{3-9} 
\multicolumn{2}{c|}{}                                                                                                                       & Graph Convolution & -             & -             & -                    & $O(2mL h)$       & $O(2mL h)$       & $O(2mL h+2q n Lh)$ \\ \Xhline{1.2pt} 
\multicolumn{1}{c|}{\multirow{4}{*}{\textbf{Training}}} & \multicolumn{1}{c|}{\multirow{3}{*}{One Batch}} & $t_{\text{conv}}$: Graph Convolution & $O(2mLd)$       & $O(2mLd)$       & $O(2mLd+2q n Ld)$      & $O(3Bhd)$        & $O(3Bhd)$        & $O(3Bhd+n_Bhd)$          \\ \cline{3-9} 
\multicolumn{1}{c|}{}                          & \multicolumn{1}{c|}{}                                                                      & $t_{\text{bpr}}$: BPR Loss                                                     & $O(2Bd)$        & $O(2Bd)$        & $O(2Bd)$               & $O(2Bd)$         & $O(2Bd)$         & $O(2Bd)$           \\ \cline{3-9} 
\multicolumn{1}{c|}{}                          & \multicolumn{1}{c|}{}                                                                      & $t_{\text{ssl}}$: InfoNCE Loss       & -             & -             & $O(B d+B n_B d)$      & -              & -              & $O(Bd+B n_B d)$        \\ \cline{2-9} 
\multicolumn{1}{c|}{}                          & \multicolumn{2}{c|}{Total} & \multicolumn{6}{c}{$(t_{\text{conv}}+t_{\text{bpr}}+t_{\text{ssl}})T$}   \\ \hline
\multicolumn{2}{c|}{\multirow{2}{*}{\textbf{Inference}}}                                                                                             & \begin{tabular}[c]{@{}c@{}}Graph \\ Convolution\end{tabular} & $O(2mLd)$       & $O(2mLd)$       & $O(2mLd)$ & $O(nhd)$ & $O(nhd)$  & $O(nhd)$    \\ \cline{3-9} 
\multicolumn{2}{c|}{}                                                                                                                       & \begin{tabular}[c]{@{}c@{}}Calculate \\ Scores\end{tabular}  & $O(|U||I|d)$    & $O(|U||I|d)$    & $O(|U||I|d)$           & $O(|U||I|d)$     & $O(|U||I|d)$     & $O(|U||I|d)$       \\ \Xhline{1.2pt} 
\end{tabular}
\end{table*}

\vspace{-2mm}
\subsection{Lighter-X in Polynomial-based Graph Filters}
\label{sec:lighterjgcf}
As mentioned in Section~\ref{sec:pre_graph_based_models}, polynomial-based graph collaborative filtering is formally equivalent to applying different polynomial bases to compute the aggregation weights for each convolutional layer, such as Jacobi polynomial bases used in JGCF~\cite{guo2023jgcf}. Under the Lighter-X framework, we can naturally incorporate polynomial-based methods by aggregating the propagation matrix $\Z$ using different polynomial bases. This approach leverages the representational power of varied bases while allowing the aggregations to be precomputed, thereby reducing computational complexity.

\noindent \textbf{LighterJGCF.} We use a low-rank random matrix as input features and precompute polynomial features at each level. The precomputed results are then fed into an MLP to learn the final embeddings of users and items. \eat{Therefore, the computational complexity in collaborative filtering can be effectively reduced. }Specifically, we utilize the low-rank feature matrix $\X$ and the decoupled framework introduced in Section~\ref{sec:lighterx} to reformulate Equation~\ref{equ:jgcf} into the following form:
\begin{equation}
\E_{low} = MLP(\Z) = MLP(\sum_{\ell=0}^L w_\ell \Z^{(\ell)}), \quad 
\Z^{(\ell)} = \J_\ell^{a, b}(\P) \X.
\end{equation}
Similarly, we obtain $\E_{mid} = tanh(\beta MLP(\X) - \E_{low})$. Taking a single-layer MLP as an example, the dimensionality of the model parameter matrix is $h\times d$, which is much smaller than that of original JGCF model ($n \times d$). In addition, the polynomial basis functions can be precomputed to accelerate the graph convolution process.

\vspace{-2mm}
\subsection{Lighter-X in GCL for Recommendation}
\label{sec:lightergcl}
The core of GCL for recommendation, as discussed in Section~\ref{sec:pre_graph_based_models}, involves generating a perturbed adjacency matrix $\hat{\A}$ through various data augmentation techniques. This matrix is then substituted into the embedding formula to derive the perturbed embedding. For example, LightGCL~\cite{cai2023lightgcl} uses truncated SVD to obtain $\hat{\A}$. Within the Lighter-X framework, we adopt the same precomputation approach to obtain the perturbed propagation matrix $\hat{\Z}$ and its corresponding embedding matrix. This strategy enables the simultaneous precomputation of both the perturbed and standard propagation matrices, thereby improving computational efficiency.

\noindent \textbf{LighterGCL.} Since LightGCN underlies the embedding learning in LightGCL, its parameter size is $n \times d$, identical to that of LightGCN. To reduce this scale, LighterGCL adopts LighterGCN as its backbone, producing embeddings $\E$ with a parameter size of $h\times d$, where $h \ll n$, significantly reducing model complexity compared to LightGCL. To further improve efficiency and scalability, LighterGCL precomputes the perturbation component $\hat{\Z}$ using the low-rank input matrix $\X$. This strategy eliminates the need to compute perturbations during training, which is often a major bottleneck in graph contrastive learning. Specifically, the perturbed representations $\hat{\Z}^{(\ell)}$ at each layer are computed in advance using the perturbed adjacency matrix $\hat{\P}$ and the input features $\X$. The final perturbed embeddings are obtained by aggregating the precomputed $\hat{\Z}^{(\ell)}$ and passing the result through an MLP for training:
\begin{equation}
\hat{\E} = MLP(\hat{\Z}) = MLP(\sum_{\ell=0}^L w_\ell \hat{\Z}^{(\ell)}), \quad \hat{\Z}^{(\ell)} = \hat{\P} \cdot \P^{\ell - 1} \X.
\end{equation}
where $\hat{\Z}^{(0)} = \X$. As a result, the repetitive perturbation generation required in conventional approaches is circumvented by leveraging the low-rank feature matrix and the decoupling framework in LighterGCL. This substantially reduces both the time and space complexity, making LighterGCL more suitable for large-scale graph-based recommendation scenarios.

\begin{table}[t]
\setlength{\abovecaptionskip}{1mm}
\setlength{\belowcaptionskip}{-1mm}
\caption{The statistics of datasets.}
\label{tab:dataset}
\begin{tabular}{l|l|l|l|l}
\hline
\textbf{Dataset} & \textbf{\#User} & \textbf{\#Item} & \textbf{\#Interaction} & \textbf{Sparsity} \\ \hline
LastFM        & 1,892   & 17,632 & 92,834        & 99.72\%  \\ 
MovieLens-1M  & 6,040   & 3,952  & 1,000,209     & 95.81\%  \\ 
MovieLens-20M & 138,493 & 27,278 & 20,000,263    & 99.47\%  \\ 
Yelp-2018     & 31,668  & 38,048 & 1,561,406     & 99.87\%  \\  \hline
\end{tabular}
\end{table}
\begin{table*}[t]
\centering
\small
\setlength{\abovecaptionskip}{2mm}
\setlength{\belowcaptionskip}{-4mm}
\caption{Performance comparison at public datasets, with metrics evaluated at @10.}
\label{tab:result_public_datasets}
\vspace{-2mm}
\begin{tabular}{ccccc|ccc|ccc|ccc}
\hline
\multicolumn{2}{c}{\multirow{2}{*}{\textbf{Method}}}                                         & \multicolumn{3}{c|}{\textbf{LastFM}}               & \multicolumn{3}{c|}{\textbf{MovieLens-1M}}         & \multicolumn{3}{c|}{\textbf{MovieLens-20M}} & \multicolumn{3}{c}{\textbf{Yelp2018}} \\ \cline{3-14} 
\multicolumn{2}{c}{}                                                                          & Recall          & NDCG            & \textit{\#Params} & Recall          & NDCG            & \textit{\#Params} & Recall    & NDCG      & \textit{\#Params}            & Recall  & NDCG    & \textit{\#Params} \\ \hline
\multirow{7}{*}{\textbf{\begin{tabular}[c]{@{}c@{}}Standard\\ Models\end{tabular}}} & BPR         & 0.1699          & 0.1632          & \textit{2.50M}                         & 0.1658          & 0.2583          & \textit{1.25M}                         & 0.1757          & 0.2207          & \textit{21.15M}                        & 0.0452          & 0.0355          & \textit{4.46M}    \\
                                                               & NeuMF       & 0.1633          & 0.1556          & \textit{2.50M}                         & 0.1416          & 0.2239          & \textit{1.25M}                         & 0.1645          & 0.1965          & \textit{21.15M}                        & 0.0313          & 0.0235          & \textit{4.46M}    \\
                                                               & NGCF        & 0.1809          & 0.1772          & \textit{2.53M}                         & 0.1462          & 0.2413          & \textit{1.28M}                         & 0.2027          & 0.2636          & \textit{21.18M}                        & 0.0459          & 0.0364          & \textit{4.49M}    \\
                                                               & DGCF        & 0.1876          & 0.1802          & \textit{2.50M}                         & 0.1783          & 0.2700            & \textit{1.25M}                         & OOM             & OOM             & \textit{21.15M}                        & 0.0527          & 0.0419          & \textit{4.46M}    \\
                                                               & RGCF        & 0.1959          & 0.1904          & \textit{2.50M}                         & \textbf{0.1909}          & 0.2774          & \textit{1.25M}                         & OOM             & OOM             & \textit{21.15M}                        & 0.0633          & 0.0503          & \textit{4.46M}    \\
                                                               & DirectAU    & 0.1771          & 0.1657          & \textit{2.50M}                         & 0.1569          & 0.2087          & \textit{1.25M}                         & 0.1098          & 0.1363          & \textit{21.15M}                        & 0.0557          & 0.0435          & \textit{4.46M}    
                                                               \\ 
                                                                                & {\revision LTGNN}       & {\revision 0.1924}          & {\revision 0.1789}          & {\revision \textit{2.50M}}                         & {\revision 0.1780}          & {\revision 0.2752}          & {\revision \textit{1.25M}}                         & {\revision 0.1303}          & {\revision 0.1743}          & {\revision \textit{21.15M}}                        & {\revision 0.0430}          & {\revision 0.0333}          & {\revision \textit{4.46M}}    \\
                                                                                & {\revision LightGODE}   & {\revision 0.2037}          & {\revision 0.1965}          & {\revision \textit{2.50M}}                         & {\revision 0.1546}          & {\revision 0.1978}          & {\revision \textit{1.25M}}                         & {\revision 0.1843}          & {\revision 0.2293}          & {\revision \textit{21.15M}}                        & {\revision 0.0585}          & {\revision 0.0468}          & {\revision \textit{4.46M}}    \\

                                                               & SVD-GCN     & 0.1688          & 0.162           & \textit{0.02M}                         & 0.1598          & 0.2484          & \textit{0.02M}                         & -               & -               & \textit{-}                             & 0.0508          & 0.0402          & \textit{0.01M}    \\
                                                               \hline
\multirow{3}{*}{\textbf{\begin{tabular}[c]{@{}c@{}}Base\\ Models\end{tabular}}} & LightGCN    & 0.1952          & 0.1878          & \textit{2.50M} & 0.1688          & 0.2650           & \textit{1.25M} & 0.2129    & 0.2730     & \textit{21.15M}    & 0.0560   & 0.0450   & \textit{4.46M}  \\
                                                                                & JGCF        & 0.2054          & 0.1971          & \textit{2.50M} & 0.1863    & {\ul 0.2823}    & \textit{1.25M} & {\ul 0.2185}    & {\ul 0.2804}    & \textit{21.15M}    & {\ul 0.0687}   & \textbf{0.0556}   & \textit{4.46M}  \\
                                                                                & LightGCL    & 0.2050           & {\ul 0.2018}    & \textit{2.50M} & 0.1592          & 0.2539          & \textit{1.25M} & 0.1172    & 0.1578    & \textit{21.15M}    & 0.0617   & 0.0496   & \textit{4.46M}  \\ \hline
\multirow{3}{*}{\textbf{Lighter-X}}                                             & LighterGCN  & 0.1946          & 0.1882          & \textit{0.40M} & 0.1818          & 0.2731          & \textit{0.19M} & 0.2108    & 0.2780     & \textit{1.70M}      & 0.0566   & 0.0451   & \textit{0.17M}  \\
                                                                                & LighterJGCF & \textbf{0.2095} & 0.1952          & \textit{0.40M}    & {\ul 0.1883} & \textbf{0.2839} & \textit{0.19M}    & \textbf{0.2268}     &   \textbf{0.2882}   &      \textit{1.70M}     & \textbf{0.0694} & {\ul 0.0538} & \textit{0.17M} \\
                                                                                & LighterGCL  & {\ul 0.2059}    & \textbf{0.2021} & \textit{0.40M}    & 0.1753          & 0.2642          & \textit{0.19M}    &  0.1688    &  0.2217    &      \textit{1.70M}     & 0.0627 & 0.0497 & \textit{0.17M}          \\ \hline
\end{tabular}
\eat{\parbox{\textwidth}{\raggedright \shepherd{\typofootnote}}}
\end{table*}
\begin{figure*}[t]
\setlength{\abovecaptionskip}{1mm}
\setlength{\belowcaptionskip}{-4mm}
	\begin{small}
		\centering
		\vspace{-2mm}
		\begin{tabular}{cccc}
              \multicolumn{4}{c}{\includegraphics[height=6.9mm]{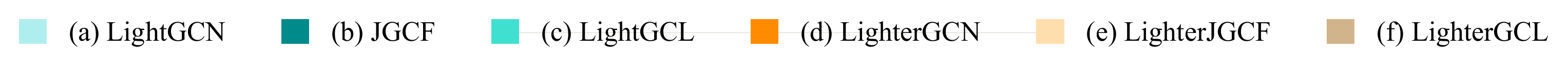}} \vspace{-2mm} \\
			 \hspace{-2mm}\includegraphics[height=32mm]{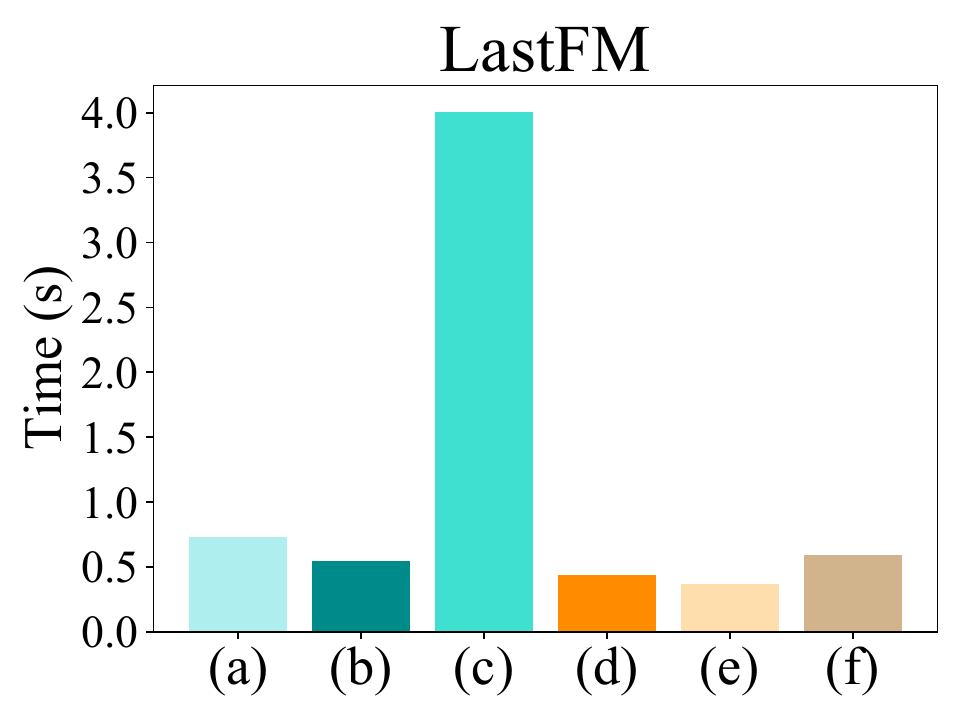} &
			 \hspace{-3mm}\includegraphics[height=32mm]{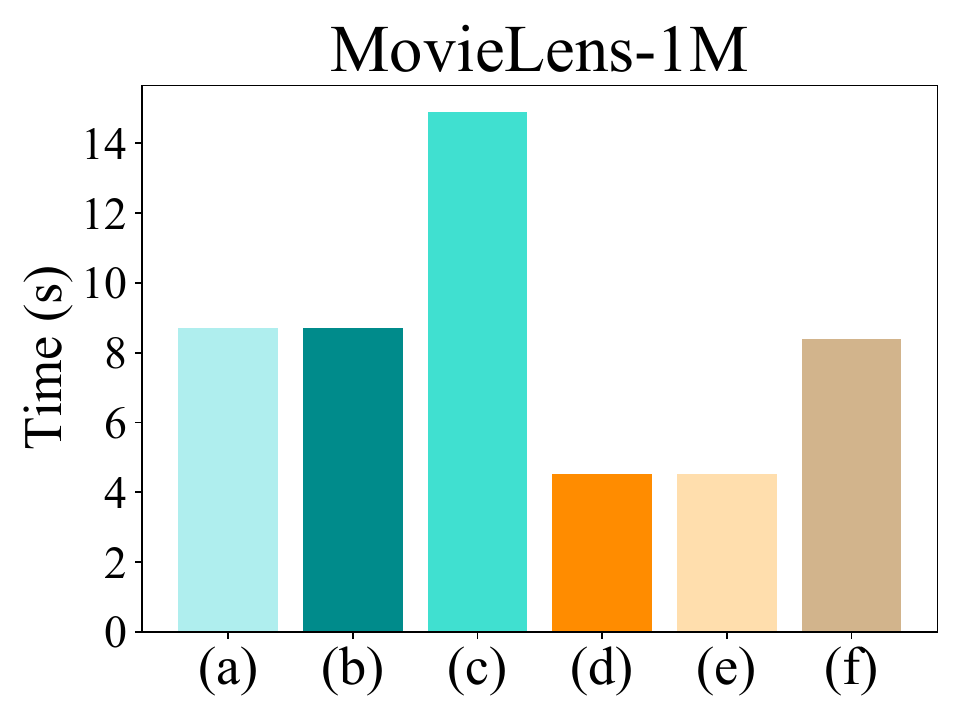} &
			 \hspace{-3mm}\includegraphics[height=32mm]{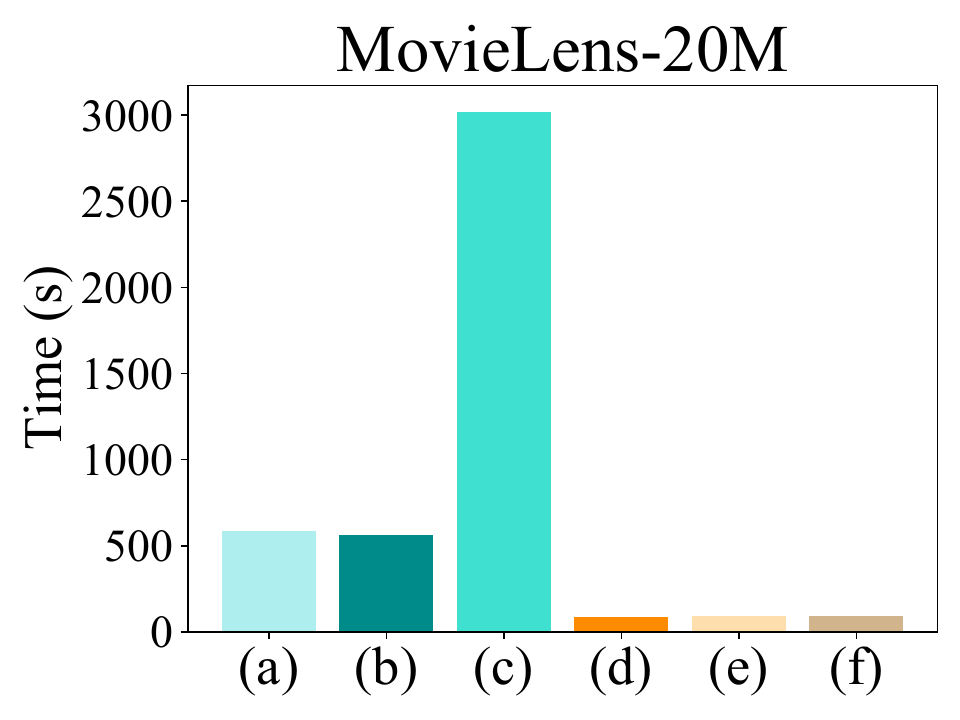} &
			 \hspace{-3mm}\includegraphics[height=32mm]{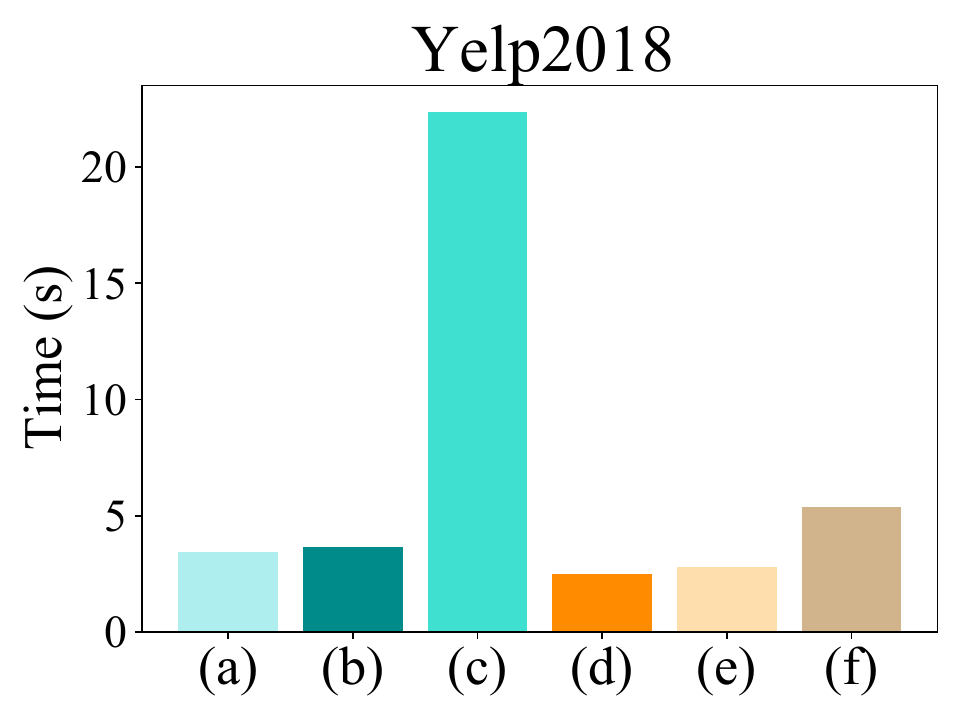}
		\end{tabular}
		\vspace{-2mm}
		\caption{Comparison of training time per epoch.}
		\label{fig:time_cost_per_epoch}
	\end{small}
\end{figure*}

\vspace{-3mm}
\subsection{Analysis}
\label{sec:analysis}
GNN-based recommendation models typically incur significant computational costs due to the need to repeatedly perform convolution operations on the entire graph during training. In contrast, we decouple the costly feature propagation from the training process, enabling models to precompute these convolution operations. This avoids redundant computations throughout training and significantly improves efficiency. Specifically, Lighter-X models only perform graph convolution during the preprocessing stage, and it only needs to be performed \textbf{once}. In contrast, baseline methods must repeat the convolution over the entire graph in \textbf{each training batch}. As shown in Table~\ref{tab:complexity}, we compare preprocessing cost, per-batch training complexity, total training complexity, and inference complexity between Lighter-X and baseline models. Due to space constraints, the detailed derivation is deferred to the technical report\eat{~\cite{technical_report}}. The results demonstrate that Lighter-X retains the theoretical advantages of its base models while substantially improving training efficiency across various applications. 

%% file: sections/experiments.tex
\section{Experiments}
\label{sec:exps}
\subsection{Experimental Setup}
\textbf{Datasets.} We conduct experiments on four datasets. (1) \textbf{LastFM} contains the listening history of users on the Last.fm online music system. (2) \textbf{MovieLens-1M} and (3) \textbf{MovieLens-20M} contain movie rating data from the MovieLens website, with each record reflecting a user's rating for a particular movie. (4) \textbf{Yelp2018} is collected from users' reviews of merchants on Yelp\footnote{\url{https://www.yelp.com/}}.  

\noindent \textbf{Baselines.}
We consider three representative models LightGCN~\cite{he2020lightgcn}, JGCF~\cite{guo2023jgcf} and LightGCL~\cite{cai2023lightgcl} as important baselines and conduct a comprehensive comparison of their performance and training efficiency against Lighter-X. Furthermore, we evaluate our models against other recommendation systems, including BPR~\cite{rendle2012bpr}, NeuMF\cite{he2017neumf}, NGCF~\cite{wang2019ngcf}, DGCF~\cite{wang2020dgcf}, RGCF~\cite{tian2022rgcf},  DirectAU~\cite{wang2022directau}, {\revision LTGNN~\cite{zhang2024ltgnn}, LightGODE~\cite{zhang2024lightgode}}, and SVD-GCN~\cite{peng2022svdgcn}, which also aims to reduce parameter counts in recommendation models. 

\vspace{-2mm}
\subsection{Experiments on Public Datasets}
\label{sec:expeiments_public}
\textbf{Evaluation Protocols.} 
In this experiment, for each user, we randomly select 80\% and 10\% of the interactions as the training and validation sets, while the others are left for testing. We use Recall and NDCG as the evaluation metrics and have the recommender models generate a ranked list of 10 or 20 items to compare against the ground truth. Due to space limitation, we focus on Recall@10 and NDCG@10 in the main paper, while the other results are provided in the technical report\eat{~\cite{technical_report}}.

\noindent \textbf{Effectiveness.}
The overall performance of our framework compared to different base models is presented in Table~\ref{tab:result_public_datasets}. We can see, our framework can achieve comparable or even better performances than the base model across all the evaluation metrics and datasets. These results are encouraging, given that our framework uses significantly fewer parameters. This suggests that many parameters in traditional graph-based recommender models may be redundant and contribute little to performance improvement. Usually, recommender systems must process large volumes of real-time data, which demands high training efficiency. The above experiments demonstrate that our lightweight framework is well-suited to meet this requirement. Finally, our framework serves as an efficient plug-and-play strategy, which makes it more flexible and practical in real-world scenarios.

\noindent \textbf{Efficiency.}
In the above experiments, we demonstrate the effectiveness of our framework. A more significant advantage of our framework is its efficiency. 
In this experiment, we analyze the time cost of our framework in the training phase. 
To evaluate the cost, we compare our framework with different base models for training one epoch. 
As shown in Figure~\ref{fig:time_cost_per_epoch}, 
our framework can greatly reduce the time cost as compared with the base model. For example, on the MovieLens-20M dataset, the training time of Lighter-X is about 1/6 of the base model's. 
This result verifies the potential of our framework for efficient model training, which is crucial for practical recommender systems.

\noindent \textbf{Performance comparison with other models.} We also compare the proposed Lighter-X method against other leading recommendation algorithms on these public datasets. Among all the baselines, JGCF~\cite{guo2023jgcf} performs the best. The proposed LighterJGCF achieves superior performance across most datasets with significantly fewer parameters. Although SVD-GCN~\cite{peng2022svdgcn} also reduces the scale of parameters, it leads to substantial performance degradation. Moreover, computing SVD efficiently on large-scale graphs remains a challenging and unresolved issue. For example, on the MovieLens-20M dataset, SVD-GCN~\cite{peng2022svdgcn} fails due to its inability to complete the SVD computation.

%% file: sections/experiments_otherscenarios.tex
{\revision
\vspace{-3mm}
\section{Evaluation in Other Scenarios}
\label{sec:otherscenarios}
Beyond general recommendation, our proposed method can be adapted to other recommendation scenarios, including non-bipartite graphs (e.g., social recommendation) and context-aware recommendation. These settings introduce additional challenges, such as increased graph sparsity and the need to incorporate contextual information. In this section, we discuss how our approach can be extended to effectively address these alternative use cases.

\vspace{-2mm}
\subsection{Non-Bipartite Graphs}
Most recommendation systems are based on bipartite graphs, where interactions occur between two distinct sets, such as users and items. In contrast, non-bipartite graph recommendation systems model more complex relationships where entities belong to the same set and can have direct connections. This is particularly relevant in scenarios like social recommendation, where users interact with each other~\cite{ma2024tailfriend, song2022friendrec}, or when items have inherent relationships, such as movies in a cinematic universe~\cite{rahman2022movierec, wei2022heterogeneous}. In graph-based recommendation models, user nodes $u$ and item nodes $i$ are mathematically equivalent in the message-passing framework. As a result, models such as LightGCN~\cite{he2020lightgcn} can be directly applied to non-bipartite graphs without requiring structural modifications.

\noindent \textbf{Datasets.} To evaluate the performance of our model on non-bipartite graph recommendation, we conducted experiments on two real-world social network datasets provided by SSNet~\cite{song2022friendrec}: Pokec and LiveJournal\footnote{\url{https://snap.stanford.edu/data/index.html\#socnets}}. The statistics of these datasets are presented in Table~\ref{tab:dataset_nonbipartite}. Notably, the graphs in these datasets are larger and sparser compared to the bipartite graph used in Table~\ref{tab:dataset}.

\noindent \textbf{Effectiveness.} We assess model performance on the candidate retrieval task, where models are required to recall the positive candidate from the entire graph. We adopt Hit@100 and NDCG@300 as evaluation metrics. As shown in Table~\ref{tab:result_friendrec}, LighterGCN consistently outperforms LightGCN while using only 0.007\% to 0.2\% of the parameters, further demonstrating the efficiency and effectiveness of the proposed method. Additionally, we compare the training times of the methods, which can be found in the technical report\eat{~\cite{technical_report}}.

\begin{table}[t]
\setlength{\abovecaptionskip}{1mm}
\caption{\revision The statistics of non-bipartite graph datasets.}
\label{tab:dataset_nonbipartite}
\small
\begin{tabular}{cccc}
\toprule
\textbf{Dataset} & \textbf{\#Node} & \textbf{\#Edge} & \textbf{Density} \\ \midrule
Pokec            & 1,632,803       & 27,560,308      & 0.0021\%              \\
LiveJournal      & 4,847,571       & 62,094,395      & 0.0005\%           \\ \bottomrule   
\end{tabular}
\vspace{-2mm}
\end{table}
\begin{table}[t]
\setlength{\abovecaptionskip}{1mm}
\setlength{\belowcaptionskip}{-1mm}
\caption{\revision Performance comparison on non-bipartite graph recommendation, `+ SS' indicates applying SSNet on the base model. Hit@100 (Hit) and NDCG@300 (NDCG) are reported.}
\label{tab:result_friendrec}
\resizebox{\linewidth}{!}{\begin{tabular}{ccccccc}
\toprule
\multirow{2}{*}{\textbf{Method}} & \multicolumn{3}{c}{\textbf{Pokec}} & \multicolumn{3}{c}{\textbf{LiveJournal}} \\ \cmidrule(lr){2-4} \cmidrule(lr){5-7}
                                 & Hit          & NDCG  & \textit{\# Params}       & Hit             & NDCG  & \textit{\# Params}         \\ \midrule
LightGCN                         & 0.0654          & 0.0236          & \textit{104.50M}  & 0.0537              & 0.0240             & \textit{310.24M}  \\
LightGCN + SS                    & 0.1645          & 0.0536          & \textit{104.50M}  & 0.2604              & 0.0747             & \textit{310.24M}  \\
LighterGCN                       & 0.0754          & 0.0252          & \textit{2.93M}    & 0.2624              & 0.0822             & \textit{2.20M}    \\
LighterGCN + SS                  & \textbf{0.1706} & \textbf{0.0552} & \textit{2.94M}    & \textbf{0.2678}     & \textbf{0.0831}    & \textit{2.20M}    \\ 
\bottomrule
\end{tabular}}
\vspace{-2mm}
\end{table}
\begin{table}[t]
\setlength{\abovecaptionskip}{1mm}
\setlength{\belowcaptionskip}{-1mm}
\caption{\revision The statistics of datasets with context. $F_u$, $F_i$, $F_{(u, i)}$ represent the number of attributes for user, item and interaction, respectively.}
\label{tab:dataset_context}
\resizebox{\linewidth}{!}{\begin{tabular}{ccccccc}
\toprule
\textbf{Dataset} & \textbf{\#User} & \textbf{\#Item} & \textbf{\#Interaction} & \textbf{$F_u$} & \textbf{$F_i$} & \textbf{$F_{(u, i)}$} \\ \midrule
MovieLens-1M-C   & 6,040           & 3,952           & 1,000,209              & 3              & 2   &    2                \\
Yelp-2018-C      & 213,171         & 94,305          & 3,277,932              & 8              & 4   &    5         \\
\bottomrule
\end{tabular}}
\vspace{-2mm}
\end{table}
\begin{table}[t]
\setlength{\abovecaptionskip}{1mm}
\setlength{\belowcaptionskip}{-1mm}
\caption{\revision Performance comparison on context-aware recommendation, with metrics evaluated at @10.}
\label{tab:result_context}
\resizebox{\linewidth}{!}{\begin{tabular}{ccccccc}
\toprule
\multirow{2}{*}{\textbf{Method}} & \multicolumn{3}{c}{\textbf{MovieLens-1M-C}} & \multicolumn{3}{c}{\textbf{Yelp2018-C}} \\ \cmidrule(lr){2-4} \cmidrule(lr){5-7}
                        & Recall   & NDCG  & \textit{\#Params}  & Recall & NDCG & \textit{\#Params} \\ \midrule
LightGCNC                & 0.1784      & 0.2713   & \textit{1.28M}     & 0.0310    & 0.0170  & \textit{39.53M}    \\
LighterGCNC              & 0.1821      & 0.2834   & \textit{0.20M}     & 0.0382    & 0.0213  & \textit{1.04M}    \\
\bottomrule
\end{tabular}}
\vspace{-2mm}
\end{table}

\vspace{-2mm}
\subsection{Context-Aware Recommendation}
In real-world recommendation scenarios, raw features are often extremely sparse, spanning hundreds of fields and millions of dimensions. To handle the high-dimensional and sparse nature of such contextual features, many studies adopt embedding techniques, which map categorical variables into low-dimensional dense vectors to compress representations and uncover latent semantic relationships~\cite{tian2023eulernet, wei2023fgcr, lian2018xdeepfm}. Therefore, we first encode the multi-field attributes extracted from user behavior logs (e.g., age, gender, location) and item metadata (e.g., price, historical purchase counts) as one-hot vectors. These vectors are then transformed into dense embeddings using attribute-specific embedding matrices. We concatenate these attribute embeddings with the graph-enhanced embeddings $\E$ obtained from user and item IDs and the graph structure (as described in Equations~\ref{equ:lightgcn} and~\ref{equ:lightergcn}) to form the final embedding. 

Building on this methodology, we propose context-aware variants, namely LightGCNC and LighterGCNC, and evaluate their performance on the MovieLens-1M-C and Yelp2018-C datasets. Detailed model specifications can be found in the technical report\eat{~\cite{technical_report}}. Dataset details are summarized in Table~\ref{tab:dataset_context}, where MovieLens-1M-C shares the same interaction data as MovieLens-1M in Table~\ref{tab:dataset}. For all experiments, the attribute embedding size is set to 16, and other experimental settings follow those described in Section~\ref{sec:expeiments_public}. Experimental results in Table~\ref{tab:result_context} show that LighterGCNC outperforms LightGCN while using only 0.03\%–0.16\% of the parameters. This demonstrates that our method can be naturally extended to context-aware recommendation scenarios while maintaining strong performance, further validating its generality.
}

%% file: sections/conclusion.tex
\vspace{-2mm}
\section{Conclusion}
In this paper, we address a prevalent issue in existing graph-based recommendation models: the extensive and redundant volume of parameters. We propose Lighter-X, an efficient plug-and-play strategy that effectively reduces model parameter count while retaining the theoretical advantages of the base models. By introducing compressed sensing, we achieve considerable expression capabilities with more compact parameters, significantly reducing the overall parameter count. By implementing decoupled propagation, efficiency and scalability of the proposed method are further improved. 
Empirical evaluations demonstrate that Lighter-X reduces parameter size and improves efficiency while maintaining comparable performance.

%% file: sections/appendix_notation.tex
\vspace{-2mm}
\section{Notations}
\begin{table}[h!]
\centering
\caption{Notations and the corresponding definitions.}
\label{tab:notation}
\begin{tabular}{ll}
\hline
Notation                         & Description                       \\ \hline
$U$ / $I$                              & the user / item set                    \\
$\R \in \mathbb{R}^{|U| \times |I|}$ & the interaction matrix               \\
$G$                              & the graph                         \\
$V$ / $E$                              & the vertex / edge set                    \\
$n$                                & the number of nodes, $n = |U| + |I|$             \\
$N(v)$ & the neighbor set of node $v \in V$              \\
$\A \in \mathbb{R}^{n \times n}$ & the adjacency matrix               \\
$\D \in \mathbb{R}^{n \times n}$ & the degree matrix                 \\
$\P \in \mathbb{R}^{n \times n}$                & the normalized adjacency matrix    \\
$\e_v^{(\ell)} $ & the embedding vector of node $v \in V$ at layer $\ell$               \\
$\E^{(\ell)} $ & the embedding matrix at layer $\ell$               \\
$\X \in \mathbb{R}^{n \times h}$ & the feature matrix                \\
$\sigma$                         & the nonlinear activation function \\
\hline
\end{tabular}
\vspace{-2mm}
\end{table}

%% file: sections/appendix_model.tex
\vspace{-2mm}
\section{Further Details of the Model}
\label{sec:further_model_details}
\subsection{Design of Low-rank Input Feature Matrix}
\label{sec:design_x}
Since user-item interactions in recommendation systems can be represented as a bipartite graph, the corresponding $\P$ matrix is a block matrix. Therefore, the input feature matrix $\X$ is expressed as:
\begin{equation}
\label{equ:calfeat}
\X  =\left[\begin{array}{ll}
\0_{|U|\times |U|} & \B \\
\B^\top & \0_{|I| \times |I|}
\end{array}\right]\left[\begin{array}{ll}
\0_{|U|\times |U|} & \S_2 \\
\S_1 & \0_{|I| \times |I|}
\end{array}\right],
\end{equation}
where $\0_{N\times N}$ is a zero matrix of dimension $N \times N$, $\B = \Dn_u \R \Dn_i$, $\D_u$ and $\D_i$ represent the diagonal degree matrix of users and items, respectively. $\S_1$ and $\S_2$ are random matrices of size $|U|\times h_1$ and $|I|\times h_2$, used to compress the matrices $\R$ and $\R^\top$, respectively. {\shepherd To implement this construction, we detail the generation process of the low-rank input feature matrix in Algorithm~\ref{alg:gen_feat}. The procedure generates two random projection matrices $\S_1$ and $\S_2$ based on compressed sensing principles and then assembles the feature matrix $\X$ using matrix multiplication as described above.}

To ensure sampling quality, we follow a rigorous theoretical foundation for sparse signal reconstruction based on the Restricted Isometry Property (RIP). This guides the design of suitable random matrices, such as Gaussian or Bernoulli random matrices, which have been shown both theoretically and empirically to satisfy the RIP condition:
\begin{equation}
\label{equ:rip_test}
(1-\delta)\|\p\|^2 \leq\|\S \cdot \p\|^2 \leq(1+\delta)\|\p\|^2,
\end{equation}
where $\p$ is a row in the matrix $\B$ or $\B^\top$ representing the sparse signal vector of a user $u\in U$ or item $i\in I$. In addition, the RIP constrains the dimension of the random matrix $\S$, 
which is determined by the sparsity level $r$. As introduced in Section~\ref{sec:lighterx}, the dimension $h$ is set as a function of $r$ and serves as a tunable hyperparameter in our implementation to control the size of the input feature matrix $\X$. 
Compressed sensing theory guarantees that a noise-free signal can be perfectly recovered when the sampling matrix $\S$ satisfies RIP. In practice the recovery process can be solved by the Basis Pursuit algorithm~\cite{candes2005cs_theory, candes2006cs_theory}. This guarantees that the sampled signals preserve the information from the original signal matrix, and indicates that in our formulation, the sampled signals $\B\S_1$ and $\B^\top \S_2$ fully capture the noise-free $\B$ and $\B^\top $ matrices when $\S_1$ and $\S_2$ both satisfy RIP.

\begin{algorithm}[t]
    \caption{\shepherd Generating Low-rank Input Feature Matrix}
    \label{alg:gen_feat}
    \begin{algorithmic}[1]
        \Function{\textcolor{functioncolor}{GenFeat}}{$\R, c$}
            \State Construct degree matrices $\D_U, \D_I$ for users and items
            \State Compute $\B = \D_U^{-\frac{1}{2}} \R \D_I^{-\frac{1}{2}}$
            \State Generate random matrices 
            \[
            \S_1 = \textsc{GenRandomMatrix}(|I|, |U|, c, \B, \tau)
            \]
            \[
             \ \ \S_2 = \textsc{GenRandomMatrix}(|U|, |I|, c, \B^\top, \tau)
            \]
            \State Formulate:
            \[
                \X = \left[ \begin{array}{ll}
                    \0 & \B \\ 
                    \B^\top & \0
                \end{array} \right]
                \cdot 
                \left[ \begin{array}{ll}
                    \0 & \S_2 \\ 
                    \S_1^\top & \0
                \end{array} \right]
            \]
            \State \textbf{return} $\X$
        \EndFunction

        \vspace{2mm}
        \Function{{GenRandomMatrix}}{$n, f, c, \B, \tau$}
            \State Determine average sparsity $r=\frac{|\B|_0}{f}$
            \State Compute $h = c \cdot r \log \left( \frac{n}{r} \right)$
            \If{$\tau$ is \textbf{Bernoulli}}
                \State Generate $\S=\text{RandomChoice}(\{1, -1\}, \text{size} = (h, n))$
            \Else
                \State Generate $\S = \tau(h, n)$
            \EndIf
            \State \textbf{return} $\S$
        \EndFunction
    \end{algorithmic}
\end{algorithm}
\subsection{Enhancing Efficiency with Sparse Trainable Parameters}
The primary goal of \textbf{Lighter-X} is to reduce the number of parameters in graph-based recommendation models. To achieve this, we control the sparsity of training parameters by incorporating low-rank random matrices into the optimization process. Specifically, we define the parameter matrix of Lighter-X as $\W^\prime = \S^\prime \W$, where $\S^\prime \in \mathbb{R}^{h\times n}$ is a random matrix. Therefore, the modified embedding calculation is:
\begin{equation}
\label{equ:lighter_W}
\E^\prime = \sum_{\ell=0}^{L} w_{\ell} \mathbf{\P}^\ell \X\W^\prime = \sum_{\ell=0}^{L} w_{\ell} \mathbf{\P}^\ell\X\S^\prime \W.
\end{equation}
As discussed in Section~\ref{sec:investigation_lightgcn}, $\W$ can be decomposed into trainable and fixed components, allowing the equation to be further refined as:
\begin{equation}
\E^\prime = \sum_{\ell=0}^{L} w_{\ell} \mathbf{\P}^\ell \X\S^\prime\W_\text{v} + \sum_{\ell=0}^{L} w_{\ell} \mathbf{\P}^\ell \X\S^\prime \W_\text{s} = \E^\prime_\text{v}+\E^\prime_\text{s} ,
\end{equation}
where $\E^\prime_\text{v}$ and $\E^\prime_\text{s}$ correspond to the embeddings derived from the varying (trainable) and static components of $\W$, respectively. Since the fixed parameters $\W_\text{s}$ contribute little to model training, $\E^\prime_\text{v}$ effectively serves as the active and essential embedding, sufficiently substituting for $\E_\text{v}$ as the practical and effective representation. In particular, under this formulation, the dimension of the trainable parameter $\W^\prime$ is $\mymathinlinehl{h \times d$, independent of the number of nodes $n}$. In typical graph-based recommendation models, $\W_\text{v}$ is initialized using a Gaussian random distribution. To satisfy the RIP condition, which is essential for ensuring compression quality, $\S^\prime$ is also initialized with a Gaussian distribution. Consequently, the product $\S^\prime \W$ inherently follows a Gaussian distribution. Therefore, we can omit the traditional compressed sensing procedure and directly initialize $\W^\prime$ with a Gaussian distribution.

\vspace{-2mm}
\subsection{The Jacobi polynomials}
The $\ell$-th order of Jacobi basis is defined as:
\begin{equation}
\J_\ell^{a, b} = 
\begin{cases}1, & \ell=0 \\ 
\frac{a-b}{2}+\frac{a+b+2}{2} x, & \ell=1 
\\ \left(\theta_\ell z+\theta_\ell^{\prime}\right) \J_{\ell-1}^{a, b}(x)-\theta_\ell^{\prime \prime} \J_{\ell-2}^{a, b}(x), & \ell \geq 2,
\end{cases}, 
\end{equation}
and 
\begin{equation}
\begin{aligned}
\theta_\ell & =\frac{(2 \ell+a+b)(2 \ell+a+b-1)}{2 \ell(\ell+a+b)}, \\
\theta_\ell^{\prime} & =\frac{(2 \ell+a+b-1)\left(a^2-b^2\right)}{2 \ell(\ell+a+b)(2 \ell+a+b-2)}, \\
\theta_\ell^{\prime \prime} & =\frac{(\ell+a-1)(\ell+b-1)(2 \ell+a+b)}{\ell(\ell+a+b)(2 \ell+a+b-2)},
\end{aligned}
\end{equation}
where $a>-1$ and $b>-1$ are parameters to control the signal filter. 

\begin{figure*}[t]
\centering
\setlength{\abovecaptionskip}{1mm}
\setlength{\belowcaptionskip}{-1mm}
\hspace{-2mm}\includegraphics[height=90mm]{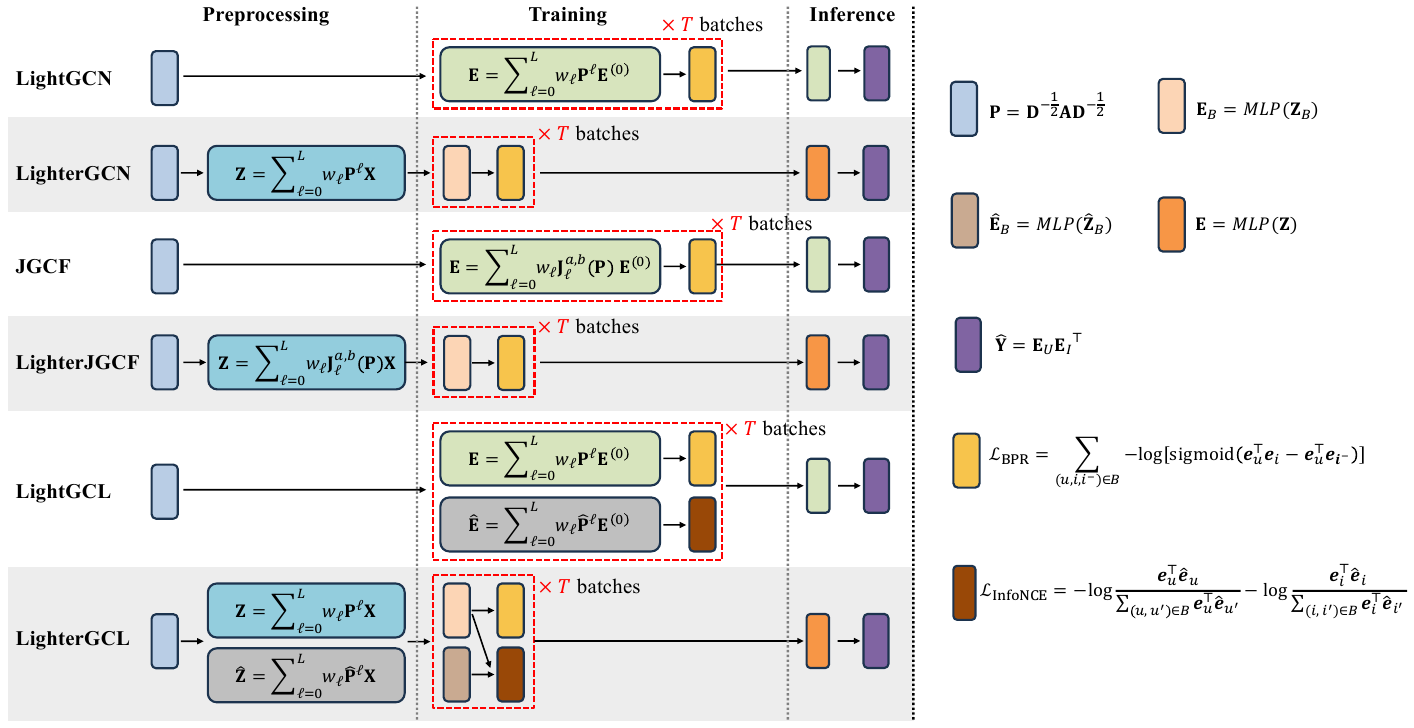}
\caption{\revision Comparison of Training Pipelines. Lighter-X models move the time-consuming graph propagation computation into the preprocessing phase, avoiding its repeated execution during training and significantly improving efficiency.}
\vspace{-2mm}
\label{fig:training_pipelines}
\end{figure*}

{\revision 
\subsection{Learning Algorithm}
Lighter-X is an efficient, plug-and-play strategy that can be seamlessly integrated with existing graph-based recommender models to reduce parameters and enhance training efficiency. Consequently, the embedding learning process in Lighter-X retains the structure of base models, except for the introduction of a randomized feature matrix and the pre-computation of graph convolution operations. 
We summarize the pseudocode of LighterGCN in Algorithm~\ref{alg:training}, and the learning procedures of LighterJGCF and LighterGCL follow the same fundamental principles as described therein. 
To avoid redundancy, we omit their pseudocode and instead summarize their training pipelines in Figure~\ref{fig:training_pipelines}, where we compare the procedures of LightGCN, JGCF, LightGCL with the proposed LighterGCN, LighterJGCF, and LighterGCL. In these pipelines, the coupled architectures of LightGCN, JGCF, and LightGCL require repeated graph convolutions during training, resulting in increased computational overhead. In contrast, our proposed LighterGCN, LighterJGCF, and LighterGCL perform the costly graph convolution operation only once during preprocessing, significantly improving training efficiency.

\subsection{Deployability in Real-World Systems}
In real-world recommendation systems, models like LightGCN~\cite{he2020lightgcn} are commonly deployed in the offline stage, where they learn final user and item embeddings by performing graph convolution and training on the user-item bipartite graph. These graph-enhanced embeddings are then combined with statistical or handcrafted features to construct input for downstream ranking models, such as those used in click-through rate (CTR) prediction. Our method follows this standard architecture and can be seamlessly integrated into existing pipelines without disrupting the overall system design.
}

%% file: sections/appendix_complexity.tex
\section{Detailed Calculation of Computational Complexity}
As shown in Table~\ref{tab:complexity}, we compare the pre-processing, per-batch training complexity, total training complexity, and inference complexity of our model against baseline models.

\begin{itemize}
    \item Compared with the original models, although Lighter-X performs graph convolution during the preprocessing stage, this operation is executed only once. In contrast, baseline methods repeatedly perform full-graph convolutions in each training batch. Therefore, Lighter-X enhances training efficiency by precomputing graph convolutions, eliminating redundant computations and significantly reducing overall computational cost.
    \item JGCF~\cite{guo2023jgcf} achieves efficient frequency decomposition and signal filtering by using Jacobi bases to approximate the graph signal filter, introducing no additional time complexity. LighterJGCF preserves this property and further improves efficiency by precomputing graph convolutions.
    \item LightGCL~\cite{cai2023lightgcl} generates more robust embeddings via graph augmentation, which increases computational complexity due to the need to compute perturbed embeddings and the InfoNCE loss during training, compared to LightGCN~\cite{he2020lightgcn} and JGCF~\cite{guo2023jgcf}. LighterGCL effectively improves the training efficiency of the model by precompleting both the graph convolution and the perturbation matrix during the preprocessing phase.
    \item In the inference stage, both baseline methods and Lighter-X perform graph convolution. Specifically, the time complexity for baseline methods is $O(2mLd)$, while that of proposed methods are $O(nhd)$, where $n$ is the number of nodes, $m$ is the number of edges, $h$ is the input feature dimension, and $d$ is the embedding size. Therefore, when the ratio $\frac{m}{n} > \frac{h}{2L}$, Lighter-X can also reduce inference time effectively.
\end{itemize}

\vspace{-2mm}
\subsection{Pre-processing}
The Pre-processing stage usually contains the Normalization operation. Lighter-X completes the graph convolution computation in this stage to avoid performing this complex computation in the training stage repeatedly. Additionally, GCL methods include data augmentations, which introduce additional complexity.

\noindent \textbf{Normalization.} This step actually computes the normalized transfer matrix $\P = \Dn \A \Dn$, which is required for all methods. In recommender systems, user-item interactions are usually modeled as an undirected bipartite graph, resulting in an adjacency matrix $\A$ with $2m$ non-zero elements, where $m$ signifies the number of user-item interactions. We adopt the sparse matrix format to store this large-scale adjacency matrix $\A$ as well as the degree matrix $\D$. Therefore, the computational complexity of performing matrix normalization is equivalent to the complexity of accessing each non-zero element in the sparse matrix, i.e. $O(2m)$.

\noindent \textbf{SVD.} LightGCL and LighterGCL employ truncated SVD for graph augmentation and precompute the SVD decomposition before training. The computing complexity is $O(qm)$, where $q$ represents the number of retained singular values. For an in-depth complexity analysis, please refer to~\cite{cai2023lightgcl, halko2011svd}.

\noindent \textbf{Graph Convolution.} LighterGCN and LighterJGCF complete the computation of $\Z=\sum_{\ell=0}^L \Z^{(\ell)}$ in this step, which involves the multiplication of the $L$-th sparse matrix $\P$ with the dense matrix $\Z^{(\ell-1)}$. Since the dimension of the matrix $\Z^{(\ell-1)}$ is $n\times h$, where $n$ denotes the number of nodes and $h$ refers to the dimension of the input features, the computation of $\P\Z^{(\ell-1)}$ requires a time complexity of $O(2mh)$. Therefore, it takes $O(2mLh)$ time to complete the computation of $\Z$. For LighterGCL, the perturbation matrix $\hat{\Z}$ needs to be computed additionally, which adopts a layer-wise perturbation approach, resulting in a complexity of $O(2qnLh)$.

\vspace{-2mm}
\subsection{Training}
All models are trained using BPR loss, which requires time to compute BPR loss. Additionally, LightGCL and LighterGCL also allocate time for computing InfoNCE loss. Moreover, models like LightGCN perform the graph convolution operation repeatedly at each batch to obtain embeddings, and LightGCL further repeats computations for perturbed embeddings. Therefore, baseline methods involve a lot of repeated computations in the training stage and require more time overall compared to Lighter-X methods.

\noindent \textbf{Graph Convolution.} Although each batch usually only involves a part of the nodes in the graph, the coupling nature of the model requires that LightGCN, JGCF, and LightGCL must complete graph convolution on the entire graph to derive the embedding of concerned nodes. For these models, the computation in this step is actually $\E=\sum_{\ell=0}^L \E^{(\ell)}$, where $\E^{(\ell)}$ denotes the embedding of the $\ell$-th layer with dimension $n\times d$, and $d$ is the embedding size. Therefore, the complexity of computing $\P\E^{(\ell)}$ is $O(2md)$. Over $L$ layers, this accumulates to a complexity of $O(2mLd)$. For LightGCL, obtaining the perturbed embedding of the $\ell$-th layer, $\hat{\E}^{(\ell)}$ from $\hat{\P}\E^{(\ell-1)}$ is essential, and proper precomputation makes the complexity of this step $O(2qnLd)$. For LighterGCN, LighterJGCF and LighterGCL, precomputation avoids the full graph convolution operation in each batch. The computational cost of this step comes from $MLP(\Z_B)$, where $\Z_B$ represents the propagation matrix of nodes involved in the batch, with dimensions of $3B\times h$ (stacking the propagation vectors of $B$ users, $B$ positive items, and $B$ negative items). Assuming that a single-layer simple MLP is used, the time to compute $\Z_B\W$ is $O(3Bhd)$, where $\W$ represents the parameter matrix with dimensions $h\times d$.

\noindent \textbf{BPR Loss.} Assume there are $B$ users in each batch. Calculating the preference scores of users for the positive and negative items requires $O(Bd)$ each, so the total complexity is $O(2Bd)$.

\noindent \textbf{InfoNCE Loss.} This step computes the comparison between positive/negative samples. Each node considers its embeddings in different views as positive samples and other nodes' embeddings as negative samples, such that the computational cost of calculating positive and negative samples is $O(Bd)$ and $O(Bn_Bd)$, respectively, where $n_B$ denotes the number of nodes within the batch.

\subsection{Inference}
For fair comparisons, we adopt the full-ranking method~\cite{he2020lightgcn, zhao2020fullrank}, assigning ranks to all candidate items that have not previously interacted with the user. Therefore, the Inference stage involves two steps: obtaining the final embedding matrix and calculating the user's preference scores for all items.

\noindent \textbf{Graph Convolution.} This computation is the same as the Graph Convolution in each batch, where the batch size is set to $n$. Therefore, the complexity of this step is $O(2mLd)$ for LightGCN, JGCF and LightGCL, and $O(nhd)$ for LighterGCN, LighterJGCF and LighterGCL.

\noindent \textbf{Calculate Scores.} Since models are evaluated using the full-ranking method, this step computes $\hat{\Y} = \E_U \E_I^\top$, where $\E_U$ and $\E_I$ denote the embedding of the user and the item, resulting in a complexity of $O(|U||I|d)$.

%% file: sections/appendix_otherscenarios.tex
{\revision 
\section{Adapting to Alternative Recommendation Scenarios}
\label{sec:apendix_otherscenarios}
In this section, we discuss the details of extending and modifying the proposed Lighter-X to effectively handle these alternative scenarios, highlighting key challenges and potential solutions.

\subsection{Application to Non-Bipartite Graph Recommendation}
Modeling recommendations in a non-bipartite structure enables capturing richer relationship patterns beyond traditional user-item interactions. In this section, we discuss how our method can be adapted to non-bipartite graph-based recommendation and analyze the challenges that arise in this setting.

In traditional recommendation methods, user-oriented and item-oriented recommendations require different handling. For example, user-oriented models rely on historical user behavior and personal preferences, while item-oriented models focus on item similarities. However, in graph-based recommendation models, user nodes $u$ and item nodes $i$ are mathematically equivalent in the message-passing framework. As a result, models such as LightGCN~\cite{he2020lightgcn} can be directly applied to non-bipartite graphs without requiring structural modifications. The final representation of a node $v \in V=\{U, I\}$ is obtained as: 
\begin{equation}
    \e_v = \sum_{\ell=0}^L w_\ell \P^\ell \e_v^{(0)} , 
\end{equation}
where $\e_v^{(0)}$ is the random initialized embeddings, which corresponds to the $v$-th row of $\E^{(0)}$ in Equation~\ref{equ:lightgcn}. To enhance the representation capability of LightGCN in friend recommendation, SSNet~\cite{song2022friendrec} introduces a self-rescaling network to improve performance. The transformation is defined as:
\begin{equation}
    \tilde{\e}_v = f(\e_v) \cdot \e , 
\end{equation}
where $f(\cdot)$ represents an additional scaling network, implemented as a two-layer MLP trained end-to-end. A Sigmoid activation function is applied to constrain the output of $f(\e_v)$ within (0, 1).

Similarly, our proposed \textbf{LighterGCN} introduces a randomized input matrix to reduce the parameter complexity of LightGCN at the source level, without modifying the message-passing equations. Consequently, it can also be directly applied to non-bipartite recommendation. It can also be directly applied to non-bipartite recommendation. The representation of node $v \in V$ is formulated as:
\begin{equation}
    \e_v = MLP(\sum_{\ell=0}^L w_\ell \P^\ell \x_v) , 
\end{equation}
where $\x_v$ refers to the $v$-th row of the low-rank input feature matrix $\X$ in Equation~\ref{equ:lightergcn}. By integrating SSNet, LighterGCN can better capture node-specific importance and refine representations, making it more effective in non-bipartite recommendation scenarios. 

\noindent \textbf{Efficiency.}  Table~\ref{tab:result_friendrec_timecompare} presents a comparison of the models in terms of per-epoch training time, the number of epochs needed for convergence, and the total training cost. The results indicate that LighterGCN significantly improves computational efficiency by reducing both the per-epoch training time and the number of epochs required for convergence. Therefore, the overall training time remains substantially lower than that of LightGCN, highlighting the advantages of LighterGCN in both scalability and efficiency for large-scale non-bipartite graph recommendation.
\begin{table}[t]
\setlength{\abovecaptionskip}{1mm}
\setlength{\belowcaptionskip}{-1mm}
\caption{\revision Training time comparison on Pokec and LiveJournal datasets (in seconds). }
\label{tab:result_friendrec_timecompare}
\resizebox{\linewidth}{!}{\begin{tabular}{c|cccc}
\toprule
\textbf{Dataset}             & \textbf{Method} & \textbf{Time/Epoch} & \textbf{\# Epochs} & \textbf{Total Time} \\ \midrule
\multirow{4}{*}{Pokec}       & LightGCN        & 63.04               & 51                 & 4049.54             \\
                             & LightGCN + SS   & 63.25               & 48                 & 3819.32             \\
                             & LighterGCN      & 1.00                & 40                 & 94.66               \\
                             & LighterGCN + SS & 1.05                & 55                 & 183.95              \\ \midrule
\multirow{4}{*}{LiveJournal} & LightGCN        & 202.21              & 78                 & 18151.20            \\
                             & LightGCN + SS   & 426.89              & 51                 & 23856.23            \\
                             & LighterGCN      & 2.43                & 29                 & 180.88              \\
                             & LighterGCN + SS & 2.64                & 32                 & 211.39     \\
\bottomrule
\end{tabular}}
\end{table}

\subsection{Application to Context-Aware Recommendation}
Compared to general recommendation, context-aware recommendation systems provide more personalized results by incorporating contextual information such as time, location, and user activities. For example, a user may prefer relaxing music at home but energetic music at the gym. Followed~\cite{tian2023eulernet}, we encode user IDs and item IDs as one-hot vectors and then obtain graph-enhanced embeddings $\E$ using LightGCN or LighterGCN, as described in Equations~\ref{equ:lightgcn} and~\ref{equ:lightergcn}. Additionally, multi-field attributes extracted from user behavior logs (e.g., age, gender, location) and item metadata (e.g., price, historical purchase counts) are also encoded as one-hot vectors. These are then transformed into dense embeddings using attribute-specific embedding matrices. By concatenating the embeddings of all relevant fields, we construct the final embedding for user $u$ as:
\begin{equation}
\h_{\mathrm{u}}=\operatorname{concat}\left(\e_u, \C_1[u], \C_2[u], \ldots \C_{F_u}[u]\right),
\end{equation}
where $\C_i \in \mathbb{R}^{j \times k}$ denotes the embedding matrix for the $i$-th attribute, $j$ represents the dimension of embedding matrix for this attribute (e.g., 2 for gender, 10 for age segments), $k$ is the attribute embedding size, and $F_u$ is the number of users' attributes. The predicted preference score of user $u$ for item $i$ is computed as $\hat{\y}_{u,i}=\h_u^\top \h_i$.

\subsection{Application to Dynamic Graph}
Real-world recommender systems often operate in dynamic environments, where user interests and interaction behaviors evolve over time, and the item pool is continuously updated. In such scenarios, recommendation models need adapt promptly to new data to maintain effectiveness. Although the proposed Lighter-X is designed for static recommendation settings, where the user-item interaction graph is assumed to remain fixed, it still shows strong potential for dynamic applications. Specifically, it can be adapted to shifting data distributions through periodic retraining (e.g., daily or hourly), enabling the model to track changes in user preferences over time. It is worth noting that the baseline models (LightGCN, JGCF, LightGCL) are also static and similarly require periodic retraining under dynamic conditions. In this context, Lighter-X offers a distinct advantage as its superior training efficiency significantly reduces the time and computational cost of each retraining cycle, enabling more frequent updates without introducing substantial latency. This makes Lighter-X a strong candidate for deployment in dynamic recommendation tasks, where maintaining a balance between recommendation accuracy and timeliness is essential for real-world online systems.

Moreover, the effectiveness of dynamic recommendation can be further enhanced by incorporating incremental learning and temporal modeling techniques. For example, temporal patterns in user behavior can be used to model the evolution of user interests~\cite{zheng2022instantgnn}. In addition, dynamic graph processing methods in GNNs, such as node state updates and edge change modeling~\cite{zheng2023decoupled}, can enable incremental updates to the local graph structure. In future work, integrating Lighter-X with these temporal modeling approaches presents a promising direction for improving its capability in real-time, dynamic environments.

}

%% file: sections/appendix_exp.tex
\section{Extended Experimental Analysis}
\label{sec:appendix_exp}
\subsection{Implementation Details} 
For all baselines and our proposed methods, we implement using RecBole~\cite{zhao2021recbole, zhao2022recbole2}, an open-source recommendation algorithm framework, and set hyperparameters based on their suggestions. All methods are optimized with Adam and initialize model parameters using the Xavier distribution. For fair comparisons, we adopt the full-ranking method~\cite{he2020lightgcn, zhao2020fullrank}, assigning ranks to all candidate items that have not previously interacted with the user. We standardize the embedding size across all methods: 64 for Yelp2018 to align with other baselines, 32 and 64 for HuaweiAds to support business processing needs, and 128 for all other datasets. For Lighter-X, we direct the configuration of the input random matrix based on RIP theory, and $c$ is turned in [1, 10]. All experiments are completed on a machine with an NVIDIA A100 GPU (80GB memory), Intel Xeon CPU (2.30 GHz) with 16 cores, and 500GB of RAM, except for the experiment on HuaweiAds which is completed on a machine with an NVIDIA Tesla V100 GPU (32GB memory), Intel Xeon CPU (2.60 GHz) with 16 cores, and 120GB of RAM.

\subsection{Hyperparameter Settings}
We employed RecBole, a unified open-source framework, to implement and reproduce various recommendation algorithms, including Lighter-X and other foundational models. To ensure a fair comparison, we set the same embedding size, $d$, for all methods and maintained consistent training parameters such as batch size. Table~\ref{tab:hyper_param} summarizes the hyperparameters of the compared methods across different datasets.

We followed the parameter recommendations of base models when setting their hyperparameters. For Lighter-X, the introduction of random matrices may necessitate adjustments to certain hyperparameters to optimize performance. Regarding hyperparameters associated with random matrices, we adjust the parameter $c$ to determine the final dimension $h$. We ensure $c \geq 1$ to meet the minimal dimensionality requirements set by the RIP test. Additionally, in denser datasets, $c$ and $h$ need to be increased to ensure that more information is retained. This method enables Lighter-X to adapt to diverse dataset densities and complexities, thereby maintaining the efficiency of dimensionality reduction while preserving crucial information for recommendation accuracy.

\begin{table*}[t]
\setlength{\abovecaptionskip}{1mm}
\setlength{\belowcaptionskip}{-1mm}
\centering
\small
\caption{Hyper-parameters of compared methods.}
\label{tab:hyper_param}
\begin{tabular}{lC{2.5cm}C{2.5cm}C{2.5cm}C{2.5cm}}
\hline
Dataset     & LastFM                                                 & MovieLens-1M                                           & MovieLens-20M                                          & Yelp2018                                               \\ \hline
LightCGN    & $d=128$                                                & $d=128$                                                & $d=128$                                                & $d=64$                                                 \\ \hline
LighterCGN  & $d=128$, $h=3096$                                      & $d=128$, $h=1348$                                      & $d=128$, $h=13160$                                     & $d=64$, $h=2632$                                     \\ \hline
JGCF        & $d=128$, $a=2$, $b=1.1$, $\beta=0.1$                   & $d=128$, $a=2$, $b=1.1$, $\beta=0.1$                   & $d=128$, $a=1$, $b=1$, $\beta=0.1$                     & $d=64$, $a=2$, $b=1$, $\beta=0.1$                      \\ \hline
LighterJGCF & $d=128$, $h=3096$, $a=2$, $b=1.2,\beta=0.3$            & $d=128$, $h=1348$, $a=1$, $b=-1$, $\beta=5$            & $d=128$, $h=13160$, $a=1$, $b=0.6$, $\beta=0.1$        & $d=64$, $h=2632$, $a=1.5$, $b=-0.5$, $\beta=0.1$      \\ \hline
LightGCL    & $d=128$, $q=5$, $\lambda_1=0.01$, $temp=0.8$           & $d=128$, $q=5$, $\lambda_1=0.01$, $temp=0.8$           & $d=128$, $q=5$, $\lambda_1=0.01$, $temp=0.8$           & $d=64$, $q=5$, $\lambda_1=0.2$, $temp=0.2$             \\ \hline
LighterGCL  & $d=128$, $h=3096$, $q=5$, $\lambda_1=0.0001$, $temp=3$ & $d=128$, $h=1348$, $q=5$, $\lambda_1=0.01$, $temp=0.8$ & $d=128$, $h=13160$, $q=5$, $\lambda_1=0.2$, $temp=0.5$ & $d=64$, $h=2632$, $q=5$, $\lambda_1=0.2$, $temp=0.5$ \\ \hline
\end{tabular}
\end{table*}
\begin{table*}[t]
\centering
\caption{Performance comparison for original and decoupled GNN models. {EqualLightGCN} denotes the decoupled GNN version corresponding to the original model {LightGCN} that employs identity matrix as the input feature, {EqualJGCF} and {EqualLightGCL} represent the equivalent decoupled versions corresponding to the original {JGCF} and {LightGCL}.}
\label{tab:equalrecs}
\begin{small}
\setlength{\abovecaptionskip}{1mm}
\setlength{\belowcaptionskip}{-1mm}
\begin{tabular}{l |C{0.9cm} C{0.9cm} C{1.1cm} C{1.1cm} |C{0.9cm} C{0.9cm} C{1.1cm} C{1.1cm} }  
\hline
\multirow{2}{*}{\textbf{Method}} & \multicolumn{4}{c|}{\textbf{MovieLens-1M}} & \multicolumn{4}{c}{\textbf{LastFM}} \\ \cline{2-9}
 & Hit@10 & MRR@10 & Recall@10 & nDCG@10 & Hit@10 & MRR@10 & Recall@10 & nDCG@10 \\ \hline
LightGCN & 0.7533 & 0.4563 & 0.1688 & 0.2650 & 0.6088 & 0.3389 & 0.1952 & 0.1878 \\
EqualLightGCN & 0.7533 & 0.4562 & 0.1689 & 0.2650 & 0.6083 & 0.3388 & 0.1951 & 0.1877  \\ \hline
JGCF & 0.7811 & 0.4822 & 0.1863 & 0.2823 & 0.6279 & 0.3513 & 0.2054 & 0.1971  \\
EqualJGCF & 0.7811 & 0.4822 & 0.1863 & 0.2823 & 0.6279 & 0.3513 & 0.2054 & 0.1971  \\ \hline
LightGCL & 0.7303 & 0.4470 & 0.1592 & 0.2539 & 0.6295 & 0.3676 & 0.2050 & 0.2018 \\
EqualLightGCL & 0.7301 & 0.4471 & 0.1593 & 0.2540  & 0.6295 & 0.3648 & 0.2064 & 0.2020 \\ \hline
\end{tabular}
\end{small}
\end{table*}
\begin{table*}[t]
\centering
\small
\setlength{\abovecaptionskip}{1mm}
\setlength{\belowcaptionskip}{-1mm}
\caption{Performance comparison at public datasets, with metrics evaluated at @20.}
\label{tab:result_public_20}
\begin{tabular}{cccc|cc|cc|cc}
\hline
\multicolumn{2}{c}{\multirow{2}{*}{\textbf{Dataset}}}                                         & \multicolumn{2}{c|}{\textbf{LastFM}} & \multicolumn{2}{c|}{\textbf{MovieLens-1M}} & \multicolumn{2}{c|}{\textbf{MovieLens-20M}} & \multicolumn{2}{c}{\textbf{Yelp2018}} \\ \cline{3-10} 
\multicolumn{2}{c}{}                                                                          & Recall            & nDCG             & Recall               & nDCG                & Recall               & nDCG                 & Recall            & nDCG              \\ \hline
\multirow{3}{*}{\textbf{\begin{tabular}[c]{@{}c@{}}Base\\ Models\end{tabular}}} & LightGCN    & 0.2730            & 0.2207           & 0.2573               & 0.2696              & 0.3071               & 0.2868               & 0.0913            & 0.0569            \\
                                                                                & JGCF        & {\ul 0.2802}      & 0.2290           & {\ul 0.2776}         & {\ul 0.2879}        & {\ul 0.3148}         & {\ul 0.2939}         & {\ul 0.1105}      & {\ul 0.0694}      \\
                                                                                & LightGCL    & 0.2793            & {\ul 0.2335}     & 0.2393               & 0.2563              & 0.1792               & 0.1669               & 0.1006            & 0.0626            \\ \hline
\multirow{3}{*}{\textbf{Lighter-X}}                                             & LighterGCN  & 0.2650            & 0.2179           & 0.2726               & 0.2797              & 0.3028               & 0.2889               & 0.0920             & 0.0571            \\
                                                                                & LighterJGCF & \textbf{0.2812}   & \textbf{0.2352}  & \textbf{0.2795}      & \textbf{0.2889}     & \textbf{0.3197}      & \textbf{0.2941}      & \textbf{0.1109}   & \textbf{0.0699}   \\
                                                                                & LighterGCL  & 0.2780            & 0.2301           & 0.2636               & 0.2699              & 0.2510               & 0.2341               & 0.1008            & 0.0632            \\ \hline
\end{tabular}
\end{table*}

\subsection{Evaluation of Equal-X}
Based on Observation~\ref{observation:key}, we found that in LightGCN, the computations for multi-layer graph convolutions can be pre-computed by utilizing the decoupled GNN model architecture. This strategy circumvents the need for computationally intensive aggregation operations at each layer. Similarly, the multi-layer graph convolution computations in JGCF~\cite{guo2023jgcf} and LightGCL~\cite{cai2023lightgcl} can also be pre-computed, utilizing the identity feature matrix. To validate this, we conducted a comparison between the original model and its equivalent decoupled GNN version (Equal-X) on two datasets: MovieLens-1M and LastFM. All equal models employ the identity matrix precomputation. As shown in Table~\ref{tab:equalrecs}, the experimental results demonstrate that decoupled Equal-X models achieve comparable performance to the original models. It means that we can further improve the efficiency of LightGCN-based recommendation models by applying precomputation techniques.

\subsection{Experiments on Public Datasets (Continued)}
As presented in Table~\ref{tab:result_public_20}, we provide the performance metrics of all methods on public datasets, evaluated at metric@20. This serves as a continuation of Table~\ref{tab:result_public_datasets}, which was not fully displayed due to page limitations.

\subsection{Online Experiments}
\textbf{Datasets. } (1) \textbf{Alimama} contains user behaviors on taobao.com platform\footnote{\url{https://tianchi.aliyun.com/dataset/56}}. We construct interaction graphs using users' purchase relationships with product categories. (2) \textbf{HuaweiAds} is a dataset containing around 3.5 million users' behaviors toward the advertisements shown on devices (mobile phone, Pad, etc). It collects a 2-hour click log from one day in 2023. Table~\ref{tab:business_dataset} summarizes the statistics of above-mentioned dataset.

\begin{table}[t]
\setlength{\abovecaptionskip}{1mm}
\setlength{\belowcaptionskip}{-1mm}
\caption{The statistics of datasets.}
\label{tab:business_dataset}
\begin{tabular}{l|l|l|l|l}
\hline
\textbf{Dataset} & \textbf{\#User} & \textbf{\#Item} & \textbf{\#Interaction} & \textbf{Sparsity} \\ \hline
Alimama       & 884,607 & 9,824  & 5,818,903     & 99.93\%  \\ 
HuaweiAds     & 1,692,592 & 25,158 & 3,504,103   & 99.99\%   \\ \hline
\end{tabular}
\vspace{-2mm}
\end{table}

\noindent \textbf{Evaluation Protocols.} 
Beyond the above experiments on public datasets, we also demonstrate the superiority of our framework in two real-world product environments including Alimama and HuaweiAds datasets.
In specific, the Alimama dataset is divided into training, validation, and testing sets in an 8:1:1 ratio based on timestamps. The HuaweiAds dataset extracts the last interaction item of each user to form the testing set, while the others are used for training.
The other settings follow the above experiments.
\begin{table}[t]
\centering
\small
\caption{Performance comparison on Alimama dataset.}
\vspace{-2mm}
\label{tab:alimama}
\begin{tabular}{cccccc}
\hline
\multirow{2}{*}{\textbf{Method}} & \multicolumn{2}{c}{\textbf{Recall@$k$}} & \multicolumn{2}{c}{\textbf{NDCG@$k$}} & \multirow{2}{*}{\textit{\#Params}} \\ \cline{2-5}
                                 & $k$=10         & $k$=20        & $k$=10            & $k$=20            &                           \\ \hline
LightGCN                         & 0.1720          & 0.1960         & 0.1538            & 0.1607            & \textit{114.49M}          \\
JGCF                             & OOM            & OOM           & OOM               & OOM               & \textit{114.49M}          \\
LightGCL                         & 0.1889         & 0.2526        & 0.1231            & 0.1413            & \textit{114.49M}          \\
LighterGCN                       & {\ul 0.2162}    & {\ul 0.2855}   & {\ul 0.1488}      & {\ul 0.1684}      & \textit{0.09M}            \\
LighterJGCF                      & \textbf{0.2241} & \textbf{0.2980} & \textbf{0.1538}   & \textbf{0.1749}   & \textit{0.09M}            \\
LighterGCL                       & 0.1967         & 0.2557        & 0.1415            & 0.1583            & \textit{0.09M}            \\ \hline
\end{tabular}
\end{table}
\begin{table}[t]
\centering
\small
\caption{Performance comparison on HuaweiAds dataset.}
\label{tab:Ads}
\vspace{-2mm}
\begin{tabular}{ccccccc}
\hline
\multirow{2}{*}{\textbf{Setting}} & \multirow{2}{*}{\textbf{Method}} & \multicolumn{4}{c}{\textbf{Recall@$k$}}                                                                        & \multirow{2}{*}{\textit{\#Params}} \\ \cline{3-6}
                                  &                                  & \multicolumn{1}{c}{$k$=1} & \multicolumn{1}{c}{$k$=3} & \multicolumn{1}{c}{$k$=5} & \multicolumn{1}{c}{$k$=10} &                                   \\ \hline
\multirow{2}{*}{$d$=32}           & LightGCN                         & 0.1218                    & 0.1724                    & 0.1974                    & 0.2352                     & \textit{54.97M}                   \\
                                  & LighterGCN                       & 0.1418                    & 0.1963                    & 0.2163                    & 0.2425                     & \textit{0.98M}                    \\ \cline{1-7} 
\multirow{2}{*}{$d$=64}           & LightGCN                         & 0.1248                    & 0.1792                    & 0.2066                    & 0.2483                     & \textit{109.94M}                  \\
                                  & LighterGCN                       & 0.1541                    & 0.2134                    & 0.2316                    & 0.2524                     & \textit{0.99M}                    \\ \hline
\end{tabular}
\end{table}
\begin{figure}[t]
\setlength{\abovecaptionskip}{1mm}
\setlength{\belowcaptionskip}{-1mm}
	\begin{small}
		\centering
		\begin{tabular}{cc}
			 \hspace{-4.3mm} \includegraphics[height=32mm]{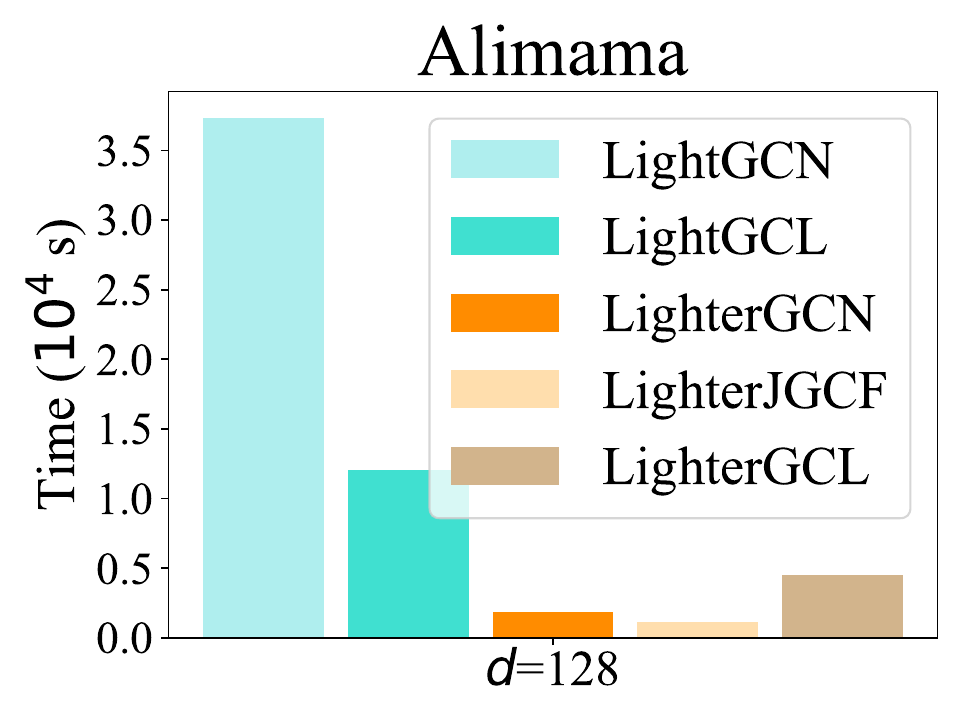} &
			 \hspace{-3.3mm} \includegraphics[height=32mm]{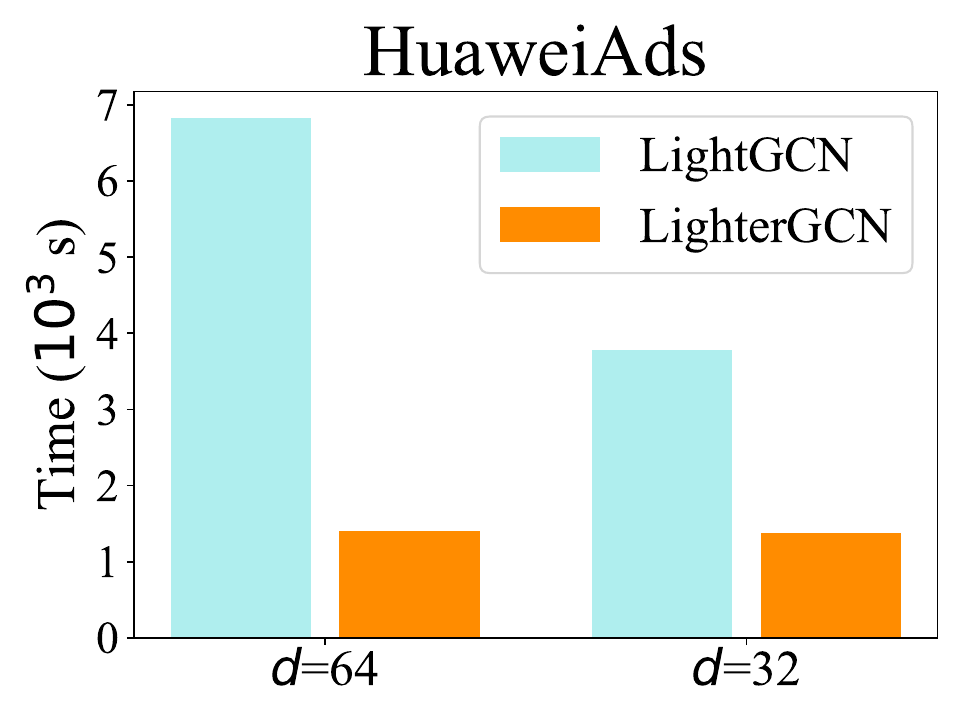} \\
		\end{tabular}
		\vspace{-2mm}
		\caption{Total running time comparison.}
		\label{fig:runing_time_real_data}
		\vspace{-5mm}
	\end{small}
\end{figure}

\noindent \textbf{Effectiveness.} Tables~\ref{tab:alimama} and~\ref{tab:Ads} show the experimental results on the Alimama and HuaweiAds datasets, respectively. We found that the model needs extra space to save the intermediate results since each $\ell$-th ($\ell \geq 2$) layer of the JGCF needs to be computed based on the embedding of the first two layers, and thus suffers from the out-of-memory (OOM) problem on the Alimama dataset. However, LighterJGCF pre-computes this computation before training, eliminating the need for repeatedly allocating additional storage space during the training phase. This enables the successful completion of the model's training. Moreover, real-world datasets are typically very sparse and encompass a vast number of users and items. As a result, baseline models require a substantial number of parameters ($n\times d$). For example, we notice that the parameter scale reaches 114.49 million for LightGCN on the Alimama dataset and 109.04 million on the HuaweiAds dataset ($d$=64). However, with just 0.09 million parameters on the Alimama dataset, which is only 0.8\% of the base model's parameters, Lighter-X achieves even better performance. Similarly, on the HuaweiAds dataset, \eat{we mainly compare LightGCN against LighterGCN. The results demonstrate that }LighterGCN attains superior performance while using just 1-1.7\% of the parameter quantity of LightGCN, which agrees with the above experiments.

\noindent \textbf{Efficiency.} In Figure~\ref{fig:runing_time_real_data}, we compare the running time of different models based on the Alimama and HuaweiAds datasets, respectively. We can see that the time cost of the base model is significantly lowered by applying our framework to it. For example, on the HuaweiAds dataset, Lighter-X reduces the total runtime by about 70\% compared to the baseline of LightGCN. 
This further validates that Lighter-X can significantly accelerate training on industrial-scale datasets.
The reduced time consumption of Lighter-X enables faster iterative optimization and more frequent updates of the model, allowing it to swiftly adapt to dynamically changing user behaviors. Consequently, our Lighter-X model exhibits superior operational and maintenance (O\&M) efficiency in industrial-scale recommender systems deployed in real-world scenarios.

{\revision
\vspace{-2mm}
\subsection{Ablation Study}
\label{sec:ablation_study}
To investigate the effectiveness of introduced randomized input features, we conduct an ablation study aimed at answering the question: Can we design suitable input feature matrices that allow the model to reduce the number of parameters and preserve performance at comparable levels? As mentioned in Section~\ref{sec:observation}, LightGCN is equivalent to LighterGCN when the input feature is an identity matrix. To reduce the dimensionality of the learnable matrix $\W$, i.e., the parameters of model, LighterGCN replaces the identity matrix features with a random matrix with dimension $n\times h$, $\X=\P\S$, where $\P$ is the normalized adjacency matrix, and $\S$ is a random matrix with dimension $n\times h$, and $h \ll n$. In order to pass the RIP test (Equation~\ref{equ:rip_test}), the random matrix $\S$ is usually generated from a Gaussian or Bernoulli distribution, and the dimension $h$ should be set according to the sparsity of the data. For simplicity, we let $\S_1$ and $\S_2$ in Equation~\ref{equ:calfeat} be obtained in the same way.

In this section, empirically examine the impact of data distribution, the dimensionality $h$ of the random matrix, and decoupled propagation on model performance. All experiments are conducted on the MovieLens-1M dataset, following the same basic experimental settings as described in Section~\ref{sec:expeiments_public}.

\noindent \textbf{Impact of \textit{data distribution}.} To examine how random matrix initialization affects performance, we developed four variants: LighterGCN-u, LighterGCN-o, LighterGCN-g, and LighterGCN-b, where the random matrix $\S$ is constructed using a uniform distribution, QR-based orthogonal projection, Gaussian distribution, and Bernoulli distribution, respectively. As shown in Table~\ref{tab:result_impact_distribution}, LighterGCN-g and LighterGCN-b consistently outperform LighterGCN-u. Although LighterGCN-o achieves better performance than LighterGCN-g at $k = 20$, it still lags behind LighterGCN-b overall. Moreover, the added complexity of QR decomposition limits its applicability on large-scale datasets. 
This reflects the widespread use of Gaussian and Bernoulli distributions in compressed sensing, due to their high likelihood of satisfying the Restricted Isometry Property (RIP), their strong universality, and their ease of generation and analysis in both theoretical and practical contexts.

\begin{table}[t]
\centering
\setlength{\abovecaptionskip}{1mm}
\setlength{\belowcaptionskip}{-1mm}
\caption{\revision The impact of data distributions.}
\label{tab:result_impact_distribution}
\resizebox{\linewidth}{!}{\begin{tabular}{ccccc}
\hline
Method        & Recall@10 & Recall@20 & NDCG@10 & NDCG@20 \\ \hline
LighterGCN-u & 0.0737    & 0.1222    & 0.1288  & 0.1308  \\
LighterGCN-o & 0.1771    & 0.2667    & 0.2723  & 0.2774  \\
LighterGCN-g & 0.1797    & 0.2653    & 0.2735  & 0.2764  \\
LighterGCN-b & 0.1818    & 0.2726    & 0.2731  & 0.2797  \\ \hline
\end{tabular}}
\end{table}

\noindent \textbf{Impact of $h$.} The dimension $h$ of the random matrix is depended on the sparsity of the data. Data in recommender systems are generally sparse, as shown in Table~\ref{tab:dataset}, and the sparsity of the interaction matrix $\R$ is usually no more than 5\%. For the purpose of reducing the number of model parameters, we want $h\ll n$ while ensuring the quality of random sampling according to Equation~\ref{equ:dim_of_random_matrix}. For sparsity $r$, we take the $k$ quantile of the user/item degree distribution in the dataset as the sparsity of the matrix $\R$. Then we turn $c$ in the range of 1 to 10. As shown in Figure~\ref{fig:impact_c_k}, larger values of $k$ and $c$ indicate a larger input feature dimension $h$, which leads to a larger number of parameters and usually implies more expressive power. However, due to the introduction of more noise, the performance does not improve by leaps and bounds. Nonetheless, this provides us with more space to trade off accuracy and computational efficiency based on practical needs.

\begin{figure}[t]
\setlength{\abovecaptionskip}{1mm}
\setlength{\belowcaptionskip}{-1mm}
	\begin{small}
		\centering
		\begin{tabular}{cc}
			 \hspace{-4.3mm} \includegraphics[height=32mm]{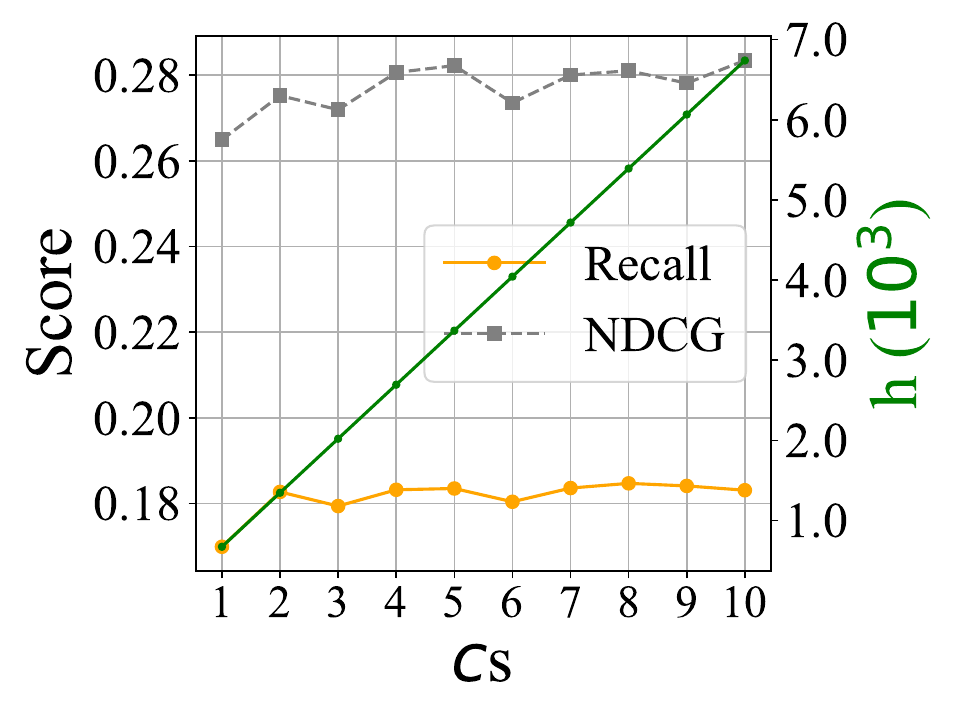} &
			 \hspace{-3.3mm} \includegraphics[height=32mm]{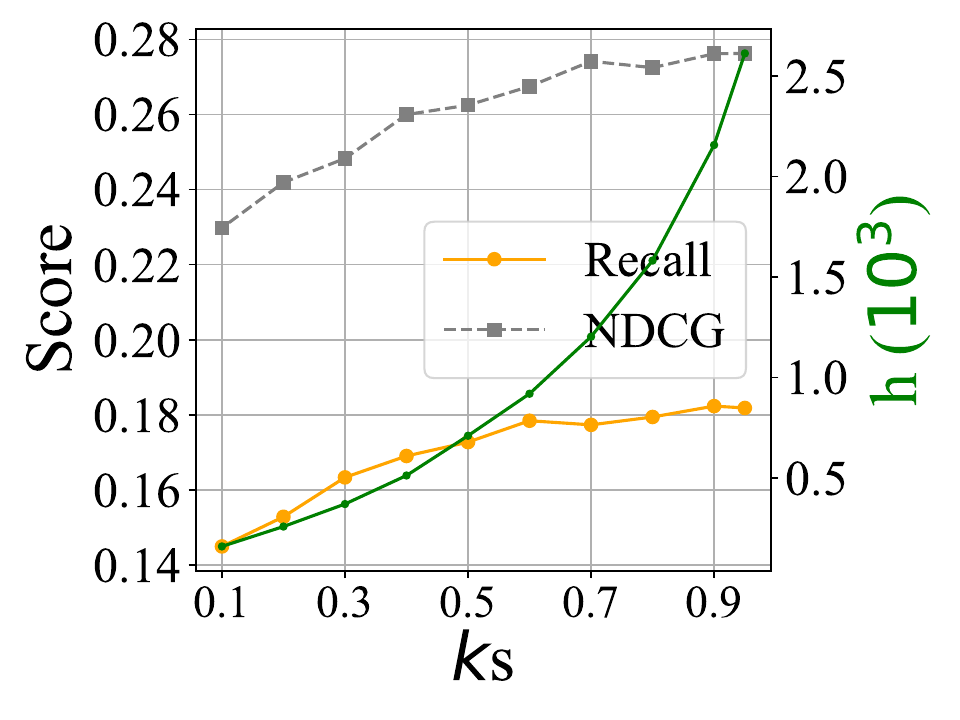} \\
		\end{tabular}
		\vspace{-4mm}
		\caption{\revision The impact of $h$.}
		\label{fig:impact_c_k}
	\end{small}
\end{figure}
\begin{table}[t]
\centering
\setlength{\abovecaptionskip}{1mm}
\setlength{\belowcaptionskip}{-1mm}
\caption{\revision The impact of decoupling and dimension reduction.}
\label{tab:ablation_decoupled}
\resizebox{\linewidth}{!}{\begin{tabular}{ccccc}
\hline
Model             & Recall@10 & NDCG@10 & \#Params & Time/Epoch \\  \hline
LightGCN          & 0.1688    & 0.265   & 1.25M    & 8.27 s         \\
EqualGCN          & 0.1689    & 0.265   & 1.25M    & 4.15 s         \\
LighterCoupledGCN & 0.1816    & 0.2731  & 0.19M    & 6.43 s          \\
LighterGCN        & 0.1818    & 0.2731  & 0.19M    & 4.12 s  \\  \hline    
\end{tabular}}
\end{table}

\noindent \textbf{Impact of the \textit{decoupled propagation}.} 
To verify the impact of the decoupled propagation, we develop two variants of LighterGCN for a comparative analysis: 
\begin{itemize}
    \item \textbf{EqualGCN} leverages the decoupled framework without incorporating random matrices for dimensionality reduction. Specifically, it utilizes the identity matrix to pre-compute the graph representation matrix $\mathbf{Z}=\sum_{\ell=0}^L \mathbf{P}^\ell \mathbf{X}$, where $\mathbf{X}=\mathbf{I}$, then it employs an MLP for subsequent training stages.
    \item \textbf{LighterCoupledGCN} integrates random matrices for dimensionality reduction but maintains the coupled structure typical of traditional models, where it recalculates $\mathbf{E} = \sum_{\ell=0}^L \mathbf{P}^\ell (\mathbf{X}\mathbf{W})$ in each training iteration.
\end{itemize}

As shown in Table~\ref{tab:ablation_decoupled}, the results indicate that EqualGCN, despite having a parameter count similar to that of traditional LightGCN, offers improved training efficiency due to its decoupled framework. Conversely, LighterCoupledGCN, while benefiting from a reduced parameter volume, does not achieve similar efficiencies owing to its retained coupled structure. These findings underscore the critical roles that both the decoupled framework and dimensionality reduction play within the proposed Lighter-X.}